\documentclass{article} % For LaTeX2e
\usepackage{iclr2027_conference,times}

\usepackage{amsmath,amsfonts,bm}

\def\eqref#1{equation~\ref{#1}}
\def\1{\bm{1}}

\DeclareMathAlphabet{\mathsfit}{\encodingdefault}{\sfdefault}{m}{sl}
\SetMathAlphabet{\mathsfit}{bold}{\encodingdefault}{\sfdefault}{bx}{n}

\usepackage{hyperref}
\usepackage{url}
\usepackage{graphicx}
\usepackage[table]{xcolor}
\usepackage{booktabs}
\usepackage{capt-of}
\usepackage{pdflscape}

\title{SimpleMemVLA: A Simple but Effective Native-Video Memory for Vision-Language-Action Models}

\author{
Cheng Yin\textsuperscript{1,8} \quad
Wang Xu\textsuperscript{3} \quad
Junpeng Yang\textsuperscript{6} \quad
Sikyuen Tam\textsuperscript{3,6} \quad
Hanyu Liu\textsuperscript{5} \quad
Yuan Yao\textsuperscript{4} \\
\textbf{\;Xiangrui Zeng\textsuperscript{1*} \quad
Junbo Cui\textsuperscript{6*} \quad
Yequan Wang \textsuperscript{7} \quad
Zhouping Yin\textsuperscript{1} \quad
Yankai Lin \textsuperscript{2,7\thanks{Corresponding author: Xiangrui Zeng, Junbo Cui and Yankai Lin}}} \\
\textsuperscript{1}Research Center for Advanced Electronics Manufacturing, MSE\thanks{MSE: School of Mechanical Science and Engineering}, HUST\thanks{HUST: Huazhong University of Science and Technology}\\
\textsuperscript{2}Gaoling School of Artificial Intelligence, Renmin University of China \\
\textsuperscript{3}Department of Computer Science and Technology, Tsinghua University \\
\textsuperscript{4}College of AI, Tsinghua University \quad
\textsuperscript{5}Peking University \quad
\textsuperscript{6}Modelbest \\
\textsuperscript{7}Beijing Academy of Artificial Intelligence \quad
\textsuperscript{8}Zhongguancun Academy, Beijing, China \\
\texttt{yinchenghust@hust.edu.cn} \quad
\texttt{cuijb2000@gmail.com} \quad
\texttt{yankailin@ruc.edu.cn} \quad
}

\iclrfinalcopy % Uncomment for camera-ready version, but NOT for submission.
\begin{document}

\maketitle

\begin{abstract}
Long-horizon manipulation is partially observable: the information needed to choose the next action may appear only in observations from minutes earlier. Existing memory mechanisms for VLAs, such as retrieval banks, learned compressors and recurrent states, must decide what to keep from the past before knowing what a future decision will require. They were motivated by the assumption that minute-scale history is too large to process directly, which no longer holds for modern VLM backbones. We propose SimpleMemVLA, a VLA without a dedicated memory module that uses the backbone's native video context directly as memory. It keeps the sampled history intact in the timestamped video format the backbone was pretrained to process, routes the evidence it finds to a standard flow-matching action head through the hidden states of a generated sub-task, and prefills the history shared by consecutive decisions during action execution, keeping latency close to that of a single-frame VLA. SimpleMemVLA achieves state-of-the-art results on four memory benchmarks without loss on general-purpose control, and with the same backbone and training setup it outperforms retrieval, compression and recurrent-state methods by a wide margin. History interventions show that the policy reads specific evidence from its past and follows edited histories without parameter updates, a visual form of in-context learning. On a physical dual-arm robot, it completes two tasks whose decisive evidence disappears before the robot acts.
\footnote{Code available at \url{https://github.com/OpenBMB/SimpleMemVLA}}
\end{abstract}

\section{Introduction}
\label{sec:intro}
As vision-language-action (VLA) models move from short tabletop skills to long-horizon tasks, partial observability becomes unavoidable~\citep{kaelbling1998}: the information needed to choose the next action—which object was revealed, where an occluded target was placed, how many times an action has already been completed—may appear only in observations from minutes earlier~\citep{masurvey,sapkotasurvey,memoryvla,hamlet}.
Most general-purpose VLAs, however, condition their actions on a single image or a sub-second observation window~\citep{rt2,octo,openvla,pi0,pi05,rdt1b,groot}.
Such policies cannot solve these tasks no matter how well they are trained, since two states with identical current observations may require different actions.

\begin{figure}[t]
\centering
\includegraphics[width=\textwidth]{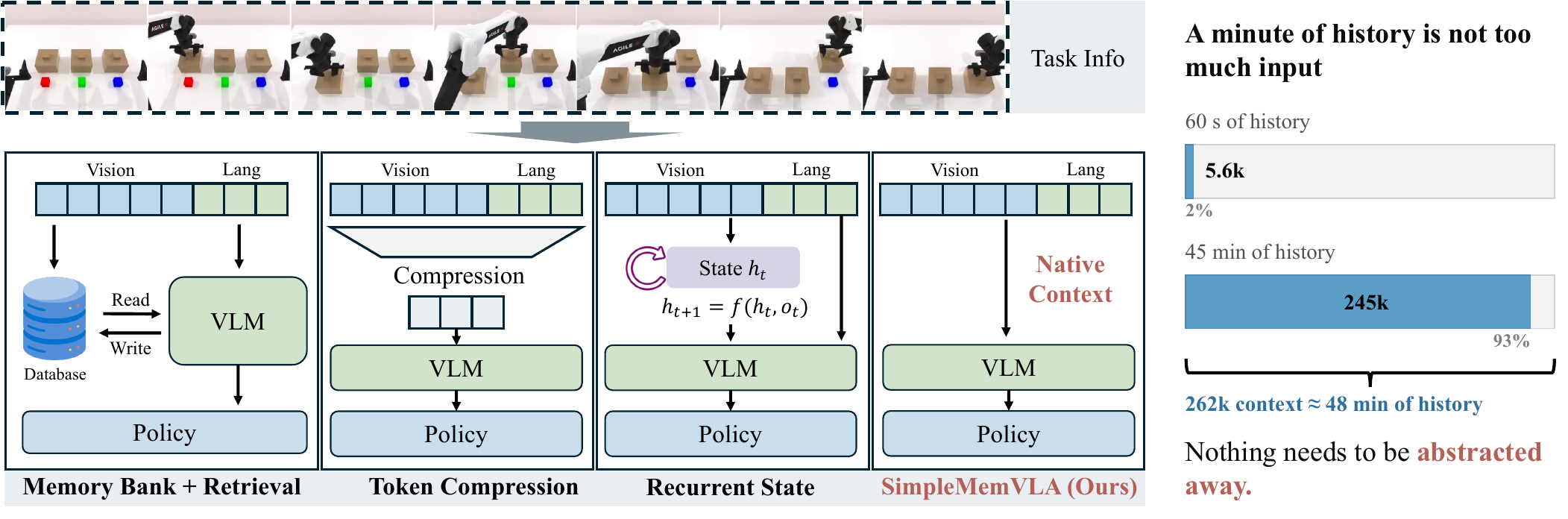}
\caption{\textbf{VLA memory mechanisms versus SimpleMemVLA.} \emph{Left:} Three prior design families insert dedicated memory machinery between the observation stream and policy, while SimpleMemVLA uses the timestamped stream directly as native context. \emph{Right:} A 60\,s history uses only 5.6k tokens of the backbone's 262k-token context window, which can hold roughly 45 minutes.}
\label{fig:overall}
\end{figure}

Existing work provides memory through dedicated mechanisms (Figure~\ref{fig:overall}): retrieval banks that select observations from an external store~\citep{mapvla,memer,echo,lin2025echovla,robomemory}, learned compressors that summarize history within a fixed token budget~\citep{memoryvla,contextvla,nativemem,hamlet}, and recurrent states that continually update a compact representation of the history~\citep{cronusvla,cherepanov2026muvlarecurrentmemorypartially,robomamba,lamem}.
Although these mechanisms differ in implementation, each must determine what remains available from the history before the needs of a future decision are known.
A relevant frame may be omitted during retrieval, visual details lost during compression, or earlier evidence overwritten by a recurrent update, and the discarded information may turn out to matter only later.
We refer to this as \emph{write-time commitment}.

These mechanisms were motivated by the assumption that minute-scale history is too large to process directly, but that assumption no longer holds.
Modern VLM backbones are pretrained to process temporally ordered video through native multimodal interfaces~\citep{qwen2vl,qwen25vl,qwen3,qwen3vl} and can read timestamped streams directly.
At the sampling rates used for manipulation, a 60\,s history occupies roughly 5.6k tokens of a 262k-token context window.
At this scale, capacity is no longer a reason to compress the past into a separate memory representation, and the open question becomes whether a policy can actually make use of a history it is simply given.

We propose \textbf{SimpleMemVLA}, a VLA without a dedicated memory module that uses the pretrained backbone's native video context directly as memory.
The idea is simple: let the policy look at what has already happened in order to decide what to do now.
We keep the sampled history intact until the current decision and present it in the timestamped video format the backbone was pretrained to process~\citep{qwen3vl}, so the backbone identifies the relevant evidence at the moment of the decision rather than relying on an earlier choice about what to retain.
This raises two questions.
First, how does the evidence found in a long history drive the action head, which conditions on a short token sequence rather than thousands of visual tokens?
We route it through a single narrow channel: the backbone generates the current textual sub-task, and the contextual hidden states and token embeddings of that span are the only path from history to a standard flow-matching action head~\citep{lipman,pi0,pi05,groot}.
Second, can a minute-scale prompt be processed within the real-time budget of a control loop?
Consecutive decisions share most of their visual history, so we prefill the shared prefix while the robot executes the current action chunk and reuse it at the next decision, which brings decision latency to 0.68\,s, close to a single-frame VLA, with outputs identical to full recomputation.

SimpleMemVLA sets a new state of the art on all four memory benchmarks and remains on par with the strongest reactive VLAs on general-purpose control, indicating that a long visual history can be carried without cost where memory is not required.
To separate the contribution of the memory interface from that of the backbone, we re-implement retrieval, compression and recurrent-state methods with the same backbone and training setup.
On RoboMME, native context reaches 88.3\%, whereas the strongest of the three reaches 31.5\%; even a symbolic pipeline supplied with ground-truth perception reaches only 84.1\%, suggesting that what limits these mechanisms is write-time commitment itself rather than the accuracy of what they commit.
History interventions confirm that the policy reads specific evidence from its past: masking one completed pick-and-place event lowers its inferred count by exactly one.
Replacing that evidence with video from another rollout redirects the policy to the substituted target without any parameter update, revealing that native video context gives rise to a visual form of in-context learning~\citep{brown2020,flamingo}.
On a physical dual-arm robot, SimpleMemVLA succeeds in 58.3\% and 70.0\% of trials on two tasks whose decisive evidence disappears before the robot acts, showing that native video context remains usable as memory under real perception and within a real-time control loop.

\section{Related Work}
\label{sec:related}

\subsection{Vision-Language-Action Models}
Most research on generalist VLAs has focused on improving how policies map the observations available at the current decision to actions.
This work spans two broad directions.
Research on \emph{action generation} has progressed from co-fine-tuned VLMs with discretized actions~\citep{rt2,openvla,nora} and control-specific tokenizers~\citep{pifast,openvlaoft} to continuous diffusion and flow-matching experts~\citep{diffusionpolicy,lipman,octo,rdt1b,pi0}.
Research on \emph{policy architecture and capability} has explored hierarchical or dual-system designs~\citep{groot,pi05}, spatial and trace representations~\citep{spatialvla,tracevla}, video-pretrained world models~\citep{gr2,worldvla,ctrlworld}, reasoning and interactive post-training~\citep{deepthinkvla,riptvla}, and cross-embodiment transfer~\citep{xvla,univla}.
These advances have improved both VLA capabilities and action generation, but how a policy should process minute-scale execution history remains an open question~\citep{masurvey,sapkotasurvey}.
SimpleMemVLA addresses this question by testing whether a pretrained backbone can process timestamped visual history directly through its native video channel without a dedicated memory mechanism.

\subsection{Memory Mechanisms for VLAs}
As VLAs are deployed in longer, partially observable tasks~\citep{kaelbling1998}, the information needed for a decision may be available only in earlier observations, making it increasingly important to preserve and reuse visual history~\citep{masurvey,sapkotasurvey,memoryvla,hamlet}.
Existing memory designs fall into four families, distinguished by what they preserve at write time before the requirements of a future decision are known: symbolic storage~\citep{conceptmem,chainvla,robomemory}, retrieval~\citep{memer,eventvla}, learned compression~\citep{memoryvla,contextvla,nativemem}, and recurrent state~\citep{cherepanov2026muvlarecurrentmemorypartially,cronusvla,lamem}.

\emph{Symbolic} pipelines provide the most explicit representation, parsing observations into structured stores outside the policy, such as scene graphs, concept banks, and execution states~\citep{robomme,conceptmem,chainvla}.
Because observations are written into a predefined schema, information outside its vocabulary is discarded during extraction, even with perfect perception.
\emph{Retrieval} methods preserve experience in an external store but expose only selected content to the policy, such as sub-trajectories, retrieved experiences, or event evidence~\citep{memer,eventvla}.
Fixed sampling schedules can be viewed as a degenerate case~\citep{fibvla}.
Because the retrieval index is constructed before the current query is available, frames that are not retrieved, along with their order and timestamps, remain unavailable to the policy for that decision.
\emph{Compression} methods instead map history into bounded learned representations, such as consolidated memory banks, amortized context tokens, or compressed visual features~\citep{memoryvla,contextvla,nativemem}.
Because the representation budget is fixed at observation time, the method must determine which perceptual details to retain before their relevance to a future decision is known.
\emph{Recurrent} methods maintain a bounded summary of the past as a continually updated state, using recurrent tokens, latent memories, or gated updates~\citep{cherepanov2026muvlarecurrentmemorypartially,lamem,gmp}.
At each update, the method must decide what to overwrite before future needs are known; once overwritten, that evidence can no longer be recovered from the state.
Despite these differences, all four families give the policy access to past observations through an intermediate memory interface: a symbolic store, a retrieval index, a compressed representation, or a recurrent state.
MEM~\citep{mem} combines compressed short-term visual context with long-term language summaries.
Its language-memory ablation finds that concatenated sub-task histories underperform compressed summaries, attributing the gap to repeated failed attempts that shift inference-time text away from demonstration histories.
SimpleMemVLA uses no such interface and instead presents minute-scale timestamped video history directly to the backbone, allowing native attention to select the past evidence relevant to each decision.
A similar result has been reported in streaming video understanding, where an off-the-shelf VLM given a sliding window of recent frames matches or outperforms dedicated streaming-memory methods~\citep{simplestream}.

\section{SimpleMemVLA}
\label{sec:method}

\begin{figure}[t!]
    \centering
    \includegraphics[width=0.95\textwidth]{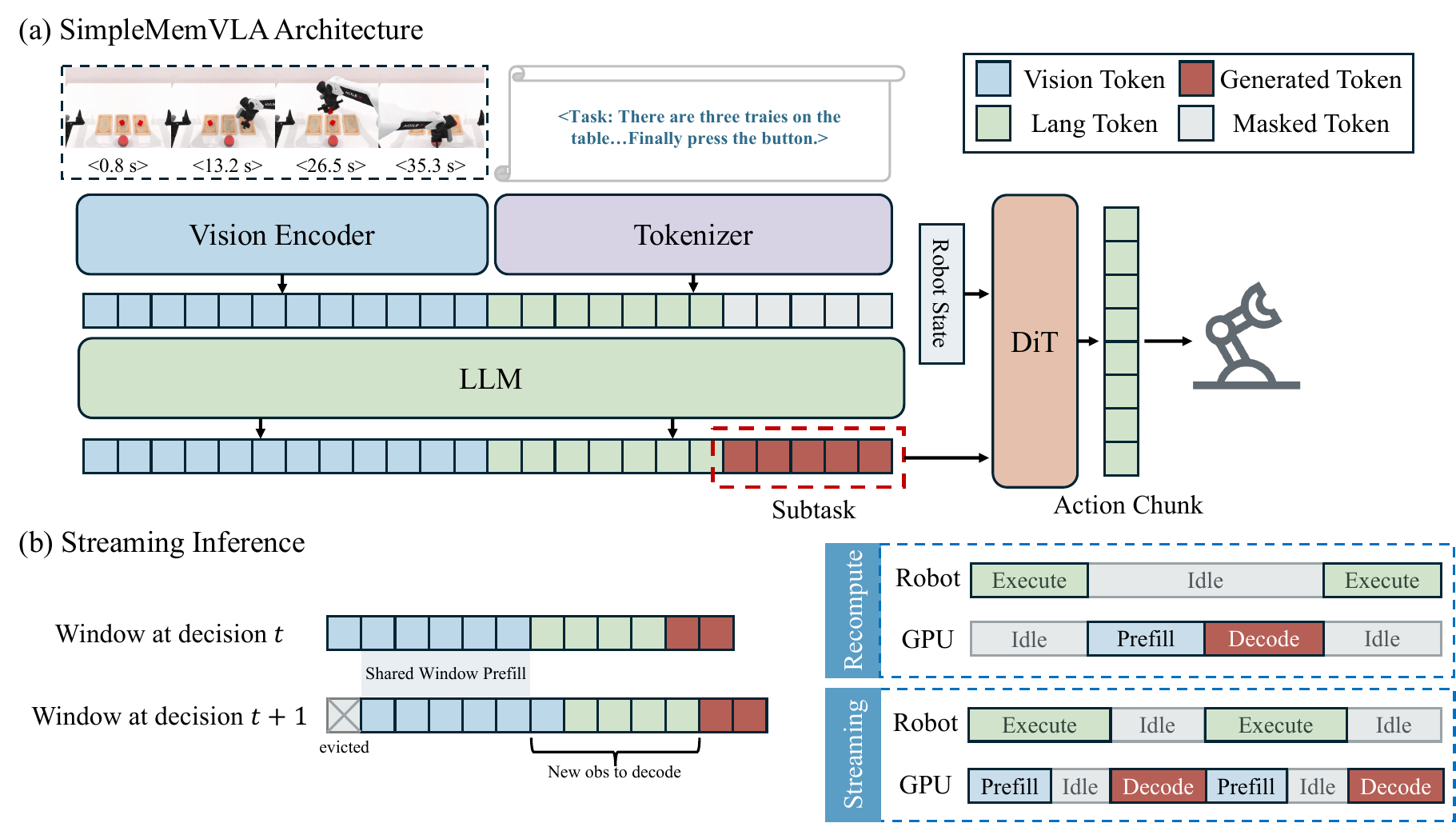}
    \caption{\textbf{SimpleMemVLA architecture and streaming inference.} \emph{(a)} The architecture uses only standard VLA components, with self-attention over plaintext-timestamped history serving as memory. \emph{(b)} Consecutive decisions differ by only one temporal patch, enabling shared-prefix prefill during action execution and reducing latency from 1.02\,s to 0.68\,s with identical outputs (Section~\ref{sec:streaming}).}
    \label{fig:arch}
\end{figure}

\subsection{Problem Setup and Overview}
\label{sec:method-setup}
We consider language-conditioned manipulation under partial observability.
At step $t$, given an instruction $\ell$, the robot receives an observation $o_t$ consisting of camera images and a proprioceptive state $q_t$, and selects an action $a_t$.
We denote the observation history together with the instruction by $h_t = (o_{\leq t}, \ell)$ and consider a history-conditioned policy $\pi(a_t \mid h_t)$.

SimpleMemVLA uses the backbone's native video context directly as memory (Figure~\ref{fig:arch}).
A window of sampled head-camera frames enters the Qwen3.5-4B backbone through its video channel, while current wrist views enter as images.
The backbone generates a one-sentence description of the current sub-task, $g_t$.
The contextual hidden states and token embeddings of this span form the only channel through which visual history reaches the flow-matching action head, which also receives current proprioception and predicts an action chunk.
We first describe how sampled history is presented as native video context (Section~\ref{sec:method-video}).
We then explain how the resulting sub-task representations drive action generation (Section~\ref{sec:method-head}) and how prefix prefill reduces decision-time latency (Section~\ref{sec:method-streaming}).

\subsection{Native Video History as Context}
\label{sec:method-video}
SimpleMemVLA represents the retained visual history as temporally ordered video with plaintext timestamps, using the backbone's native video interface~\citep{qwen3vl}.
Let $f_c$ be the native frame rate of the observation stream and $o^{\mathrm{h}}_t$ the head-camera frame at step $t$.
Rather than maintaining a separate learned memory state, SimpleMemVLA constructs, at every prediction step, a window covering the last $T_w$ seconds subsampled at a rate $f_v \ll f_c$ into at most $K = T_w f_v$ frames,
\begin{equation}
V_t \;=\; \big(o^{\mathrm{h}}_{t-(K-1)s},\, \ldots,\, o^{\mathrm{h}}_{t-s},\, o^{\mathrm{h}}_t\big), \qquad s = f_c / f_v,
\label{eq:window}
\end{equation}
where $s$ is the subsampling stride and an episode younger than $T_w$ simply yields a shorter clip.
Per suite, $T_w$ is set to cover the horizon over which its tasks leave evidence and $f_v$ is the lowest rate that does not skip decisive events, trading token budget against coverage (Table~\ref{tab:impl}).
$V_t$ enters the backbone through its \emph{video} channel, whose processor groups adjacent frames into temporal patches and prefixes each patch with the backbone's native plaintext timestamp, exactly as in video pretraining~\citep{qwen3vl}.
Standard deployment uses window-relative timestamps, labeling each patch by its offset within the active window, whereas the exploratory SWA variant in Appendix~\ref{app:swa} uses episode-absolute timestamps to keep cached patches immutable during future continual operation over unbounded input streams.
These timestamps provide an explicit temporal reference by encoding each patch's position within the window in a format the backbone already understands, allowing it to locate observed events in time relative to the current decision.

The current wrist frames $\{o^{\mathrm{w},i}_t\}_{i=1}^{W}$, where $W$ is the number of wrist cameras, enter through the \emph{image} channel without timestamps, under a single modality rule: multi-frame cameras become video and single-frame cameras become images.
The prompt builder $\Phi$ combines the history video, current wrist images and language instruction as follows:
\begin{equation}
x_t \;=\; \Phi\big(V_t,\, \{o^{\mathrm{w},i}_t\}_{i=1}^{W},\, \ell\big).
\label{eq:prompt}
\end{equation}
The prompt also includes a plain-text description of the embodiment, camera layout and timestamp convention.
The next section describes how evidence read from this context reaches the action head.

\subsection{Action Generation from Sub-task Hidden States}
\label{sec:method-head}
This section describes how information selected from the visual history reaches the action expert.
The expert accepts only a short token sequence rather than the thousands of visual tokens in the history window, so the backbone must distill the relevant information into a compact conditioning signal.
SimpleMemVLA uses the generated sub-task span as this interface, yielding a signal that is compact, directly inspectable and editable.
The backbone $f_\theta$, with parameters $\theta$, generates this span, while the DiT-style flow-matching expert $v_\phi$, with parameters $\phi$, conditions on its contextual representation and the current proprioceptive state.
We define the generated sub-task, conditioning set and conditional flow-matching objective~\citep{lipman} as:
\begin{equation}
    \begin{aligned}
    g_t &\;=\; (g_{t,1}, \ldots, g_{t,m}) \;\sim\; f_\theta(\,\cdot \mid x_t),\\
    C_t &\;=\; \big[\, e(g_{t,1}) \!\oplus\! h(g_{t,1}),\; \ldots,\; e(g_{t,m}) \!\oplus\! h(g_{t,m}),\; \psi(\bar q_t) \,\big],\\
    \mathcal{L}_{\mathrm{act}} &\;=\; \mathbb{E}_{\tau,\,\varepsilon}\, \big\| v_\phi\big(A^{\tau}, \tau \mid C_t\big) - \big(\varepsilon - \bar A_t\big) \big\|_2^2.
    \end{aligned}
\label{eq:cond}
\end{equation}
Here, $g_t$ is a one-sentence description of the robot's current sub-task, generated as an ordinary assistant response under an unmodified chat template rather than as chain-of-thought. 
The function $h(\cdot)$ returns the backbone hidden states over this response, $e(\cdot)$ its corresponding token embeddings, $\oplus$ denotes their fusion and $\psi(\bar q_t)$ is a single-token encoding of the normalized current proprioceptive state.
In the objective, $\bar A_t \in \mathbb{R}^{H \times d_a}$ denotes the normalized action chunk for the next $H$ steps, with each of its $d_a$ action dimensions z-scored using dataset statistics. 
We sample noise $\varepsilon \sim \mathcal{N}(0,I)$ and a flow time $\tau \in [0,1]$ from a distribution biased toward the noise endpoint, then construct the linear path $A^\tau=(1-\tau)\,\bar A_t+\tau\,\varepsilon$.
Because the action expert receives no prompt tokens directly, all history-dependent information needed for control must reach it through the sub-task span.

Training uses demonstrated action chunks and annotated sub-tasks $g_t^*$.
The sub-task labels are generated offline by a cloud VLM, which is given each demonstration and describes the sub-task underway at every supervision anchor (Appendix~\ref{app:impl}).
During training, the backbone processes the annotated sub-task under teacher forcing, and the action-conditioning sequence $C_t$ is constructed by substituting $g_t^*$ for $g_t$ in Equation~\ref{eq:cond}.
The same annotations supervise all three mechanism variants in Section~\ref{sec:controlled}, keeping sub-task supervision fixed in the comparison.
The joint objective is
\begin{equation}
\mathcal{L} = \lambda_{\mathrm{sub}}\mathcal{L}_{\mathrm{sub}} + \lambda_{\mathrm{act}}\mathcal{L}_{\mathrm{act}},
\label{eq:objective}
\end{equation}
where $\mathcal{L}_{\mathrm{sub}}$ is the token-level cross-entropy over the annotated answer span.

\begin{figure}[t]
    \centering
    \includegraphics[width=\textwidth]{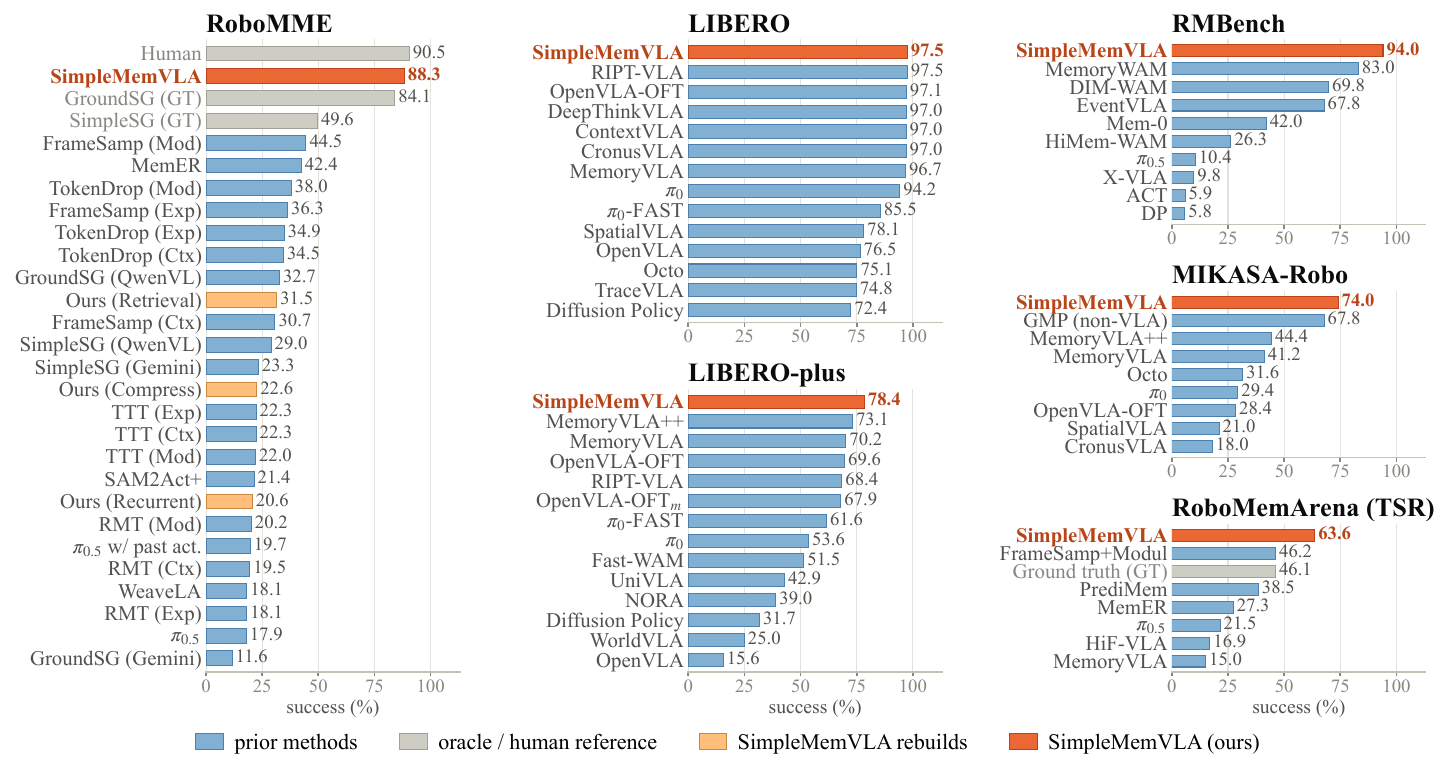}
    \caption{\textbf{SimpleMemVLA leads all memory suites and matches the best results on the general-purpose suites.} All methods from the per-suite tables are shown in score order. SimpleMemVLA uses one model per suite, while most RMBench baselines are task-specific specialists. Gray oracle and human references are excluded from ranking.}
    \label{fig:radar}
\end{figure}

At deployment, the backbone instead generates $g_t$ autoregressively from $x_t$, and the action expert uses its representations through the same conditioning rule.
In both stages, the sub-task hidden states are contextualized by the visual history and can carry task information beyond the visible wording.
Our interface ablations show that history-dependent control is carried primarily by these hidden states, while token embeddings support stable execution (Section~\ref{sec:ablations}).

The expert initializes an action chunk from Gaussian noise and integrates the learned velocity field toward a clean chunk using a small number of Euler steps.
The resulting chunk is denormalized and its first $n_e$ actions are executed.
Observations continue to be buffered at the native control rate during execution, allowing the window in Equation~\ref{eq:window} to be reconstructed exactly for the next prediction.

\subsection{Streaming Inference with Prefix Prefill}
\label{sec:method-streaming}
The final requirement is deployment efficiency: retaining minute-scale history should not place the cost of reprocessing the entire window on the critical path of every decision. 
Consecutive decisions share nearly the entire video-history prefix, so SimpleMemVLA prefills this shared prefix while the robot executes the current action chunk and stores the resulting key--value cache. 
At the next decision, the policy processes only the newly arrived temporal patch and the text instruction before decoding the next sub-task and action chunk. 
Overlapping history processing with action execution reduces decision-time latency without changing the policy output (Section~\ref{sec:streaming}).

The current scheme makes bounded, minute-scale native context practical for deployment. 
As a step toward continual inference over native video streams of unbounded duration, Appendix~\ref{app:swa} explores a variant trained with sliding-window attention (SWA), which keeps its active context and cache bounded as the history grows.

Neither the native-context memory nor its streaming implementation depends on a particular robot. 
Within the architecture, embodiment-specific choices enter only through the configuration tuple $(\mathcal{C}{\mathrm{hist}}, \mathcal{C}{\mathrm{cur}}, T_w, f_v, H, d_a)$, which specifies the history and current camera sets, window length, sampling rate, action horizon and action dimensionality. 
Moving between bimanual and single-arm platforms therefore changes this configuration rather than the memory mechanism. Appendix~\ref{app:impl} lists the concrete configuration used for each benchmark.

\section{Experiments}
\label{sec:experiments}
We organize our experiments around five questions.
First, \textbf{how effective is SimpleMemVLA?} We evaluate it on four memory-centric and two general-purpose benchmarks against published baselines (Section~\ref{sec:main-results}) and on two history-dependent
tasks using a physical dual-arm robot (Section~\ref{sec:real-world}).
Second, \textbf{does the gain come from the memory interface itself?} Holding everything else fixed, we rebuild one method from each mechanism family on our stack and compare them with native video context (Section~\ref{sec:controlled}).
Third, \textbf{does the policy actually use its visual history as memory?} Holding the current observation and policy fixed, we remove or counterfactually replace the evidence in earlier frames and measure whether the output changes (Section~\ref{sec:analysis-read}).
Fourth, \textbf{can long-context memory be deployed efficiently?} We evaluate streaming inference, which reuses the shared history across consecutive decisions instead of recomputing it (Section~\ref{sec:streaming}).
Finally, \textbf{how does each element of the recipe contribute?} We ablate history length, frame order, plaintext timestamps and the hidden-state and token-embedding inputs to the action head (Section~\ref{sec:ablations}).

\subsection{Experimental Setup}
\label{sec:setup}
We evaluate SimpleMemVLA on four memory-centric benchmarks---RMBench~\citep{rmbench}, RoboMME~\citep{robomme}, MIKASA-Robo~\citep{mikasa} and RoboMemArena~\citep{robomemarena}---alongside LIBERO~\citep{libero} and LIBERO-Plus~\citep{liberoplus} for general-purpose manipulation and zero-shot robustness, and two real-world tasks (Section~\ref{sec:real-world}).
We train one model per simulation training suite, shared across its tasks. LIBERO-Plus reuses the LIBERO model without further training.
Main-text simulation results use closed-loop evaluation on held-out seeds and streaming inference (Section~\ref{sec:streaming}).
Benchmark descriptions, evaluation protocols and baseline provenance are provided in Appendix~\ref{app:impl}, with the SWA variant evaluated separately in Appendix~\ref{app:swa}.

\begin{table}[t]
\centering
\scriptsize
\setlength{\tabcolsep}{4pt}
\caption{\textbf{RMBench} per-task success rates (\%), averaged over the nine tasks with published baselines. 
Most baselines train one specialist model per task while SimpleMemVLA is a single multi-task model evaluated at $n=100$ seeds per task in the streaming deployment of Section~\ref{sec:streaming}. 
$^{\dagger}$No prior VLA reports Place-Mat so it is excluded from all averages. Best per column in \textbf{bold}.}
\label{tab:rmbench}
\resizebox{0.99\linewidth}{!}{
\begin{tabular}{lrrrrrrrrrrrrr}
\toprule
 & \multicolumn{6}{c}{Single-memory $M(1)$} & \multicolumn{6}{c}{Multi-memory $M(n)$} & \\
\cmidrule(lr){2-7}\cmidrule(lr){8-13}
Method & \rotatebox{90}{Obs\&PU} & \rotatebox{90}{Rearr.} & \rotatebox{90}{PutBack} & \rotatebox{90}{SwapB} & \rotatebox{90}{SwapT} & \emph{Avg} & \rotatebox{90}{Battery} & \rotatebox{90}{Rank.} & \rotatebox{90}{Cover} & \rotatebox{90}{Press} & \rotatebox{90}{Place-Mat$^{\dagger}$} & \emph{Avg} & Overall \\
\midrule
\multicolumn{14}{l}{\emph{One specialist model per task}} \\
DP~\citep{diffusionpolicy} & 1 & 0 & 0 & 11 & 20 & 6.4 & 10 & 10 & 0 & 0 & -- & 5.0 & 5.8 \\
ACT~\citep{aloha} & 1 & 29 & 0 & 2 & 2 & 6.8 & 19 & 0 & 0 & 0 & -- & 4.8 & 5.9 \\
X-VLA~\citep{xvla} & 9 & 13 & 18 & 16 & 3 & 11.8 & 26 & 1 & 2 & 0 & -- & 7.3 & 9.8 \\
Mem-0~\citep{rmbench} & 4 & 89 & 90 & 67 & 14 & 52.8 & 28 & 18 & 68 & 0 & -- & 28.5 & 42.0 \\
DIM-WAM~\citep{dimwam} & 13 & 99 & 98 & 96 & \textbf{97} & 80.6 & 48 & 87 & 56 & 34 & -- & 56.3 & 69.8 \\
MemoryWAM~\citep{memorywam} & 27 & \textbf{100} & \textbf{100} & \textbf{100} & 94 & 84.2 & 41 & \textbf{100} & \textbf{98} & 87 & -- & 81.5 & 83.0 \\
\midrule
\multicolumn{14}{l}{\emph{Other published baselines}} \\
$\pi_{0.5}$~\citep{pi05} & 9 & 13 & 11 & 24 & 15 & 14.4 & 16 & 6 & 0 & 0 & -- & 5.5 & 10.4 \\
HiMem-WAM~\citep{himemwam} & 28 & 33 & 32 & 38 & 27 & 31.6 & 28 & 24 & 19 & 8 & -- & 19.8 & 26.3 \\
\midrule
\multicolumn{14}{l}{\emph{Single multi-task checkpoint}} \\
EventVLA~\citep{eventvla} & 21 & 96 & 95 & 96 & 87 & 79.0 & 35 & 81 & 97 & 3 & -- & 54.0 & 67.8 \\
\rowcolor{orange!15}
SimpleMemVLA (Ours) & \textbf{65} & \textbf{100} & \textbf{100} & \textbf{100} & 93 & \textbf{91.6} & \textbf{90} & \textbf{100} & \textbf{98} & \textbf{100} & \textbf{100} & \textbf{97.0} & \textbf{94.0} \\
\bottomrule
\end{tabular}
}
\end{table}

\subsection{Effectiveness Across Benchmarks}
\label{sec:main-results}
\label{sec:general}

\textbf{Leading performance across memory benchmarks.}
SimpleMemVLA leads all four memory benchmarks with 94.0\% on RMBench (+11.0 points), 88.3\% on RoboMME (+43.7), 74.0\% on MIKASA-Robo (+29.6) and 63.6\% task success on RoboMemArena (+17.4), relative to the strongest prior VLA baseline in each suite (Figure~\ref{fig:radar}; Tables~\ref{tab:rmbench}--\ref{tab:robomemarena}).

\begin{table}[t]
\centering
\scriptsize
\setlength{\tabcolsep}{5pt}
\caption{\textbf{RoboMME} category-level success rates (\%). AVG is over all sixteen tasks. Gray rows are reference-only and excluded from ranking. \textbf{Bold} and \underline{underline} mark ranks 1 and 2. The SimpleMemVLA variant rows re-create one mechanism family each on the otherwise unchanged SimpleMemVLA stack. Per-task results are in Table~\ref{tab:robomme-full}.}
\label{tab:robomme}
\resizebox{0.96\linewidth}{!}{
\begin{tabular}{lllrrrrr}
\toprule
Memory & Method & Integr.\ / VLM & \rotatebox{90}{Counting} & \rotatebox{90}{Permanence} & \rotatebox{90}{Reference} & \rotatebox{90}{Imitation} & AVG \\
\midrule
\rowcolor{gray!15}
Human & Human Performance & - & 88.5 & 91 & 93 & 89.5 & 90.5 \\
\midrule
\rowcolor{gray!15}
Symbolic (Oracle) & SimpleSG~\citep{robomme} & GT VLM & 82.6 & 21.6 & 32.3 & 61.9 & 49.6 \\
\rowcolor{gray!15}
Symbolic (Oracle) & GroundSG~\citep{robomme} & GT VLM & 83.9 & 93.3 & 95.2 & 64.0 & 84.1 \\
\midrule
Symbolic & SimpleSG & Gemini & 39.0 & 13.5 & 22.5 & 18.0 & 23.3 \\
Symbolic & SimpleSG & QwenVL & 44.6 & 19.6 & 25.2 & 26.6 & 29.0 \\
Symbolic & GroundSG & Gemini & 12.8 & 15.8 & 11.3 & 6.5 & 11.6 \\
Symbolic & GroundSG & QwenVL & 38.0 & 39.3 & 31.6 & 21.9 & 32.7 \\
\midrule
Perceptual & TokenDrop~\citep{robomme} & Context & 57.9 & 26.9 & 23.7 & 29.5 & 34.5 \\
Perceptual & TokenDrop & Modul & 52.3 & 26.8 & 34.7 & 38.3 & 38.0 \\
Perceptual & TokenDrop & Expert & 59.4 & 23.6 & 25.3 & 31.1 & 34.9 \\
Perceptual & FrameSamp~\citep{robomme} & Context & 50.1 & 23.3 & 20.9 & 28.3 & 30.7 \\
Perceptual & FrameSamp & Modul & 65.2 & 25.1 & 36.3 & \underline{51.4} & \underline{44.5} \\
Perceptual & FrameSamp & Expert & \underline{66.8} & 25.2 & 24.1 & 28.9 & 36.3 \\
\midrule
Recurrent & TTT~\citep{robomme} & Context & 36.1 & 21.4 & 22.0 & 9.7 & 22.3 \\
Recurrent & TTT & Modul & 34.5 & 22.2 & 20.4 & 10.7 & 22.0 \\
Recurrent & TTT & Expert & 36.0 & 22.9 & 19.7 & 10.8 & 22.3 \\
Recurrent & RMT~\citep{robomme} & Context & 32.2 & 15.4 & 19.7 & 10.6 & 19.5 \\
Recurrent & RMT & Modul & 34.1 & 15.8 & 21.1 & 9.7 & 20.2 \\
Recurrent & RMT & Expert & 34.4 & 15.7 & 14.7 & 7.8 & 18.1 \\
\midrule
Other & $\pi_{0.5}$~\citep{pi05} & - & 28.8 & 17.0 & 17.2 & 8.8 & 17.9 \\
Other & WeaveLA~\citep{weavela} & - & 31.0 & 19.5 & 12.5 & 9.5 & 18.1 \\
Other & $\pi_{0.5}$ w/ past actions~\citep{robomme} & - & 29.1 & 22.8 & 15.9 & 11.2 & 19.7 \\
Other & SAM2Act+~\citep{sam2act} & - & 35.3 & 26.0 & 16.8 & 7.3 & 21.4 \\
Other & MemER~\citep{memer} & - & 48.8 & \underline{53.2} & \underline{38.0} & 29.5 & 42.4 \\
\midrule
Ours variant & SimpleMemVLA (Retrieval) & - & 46.5 & 42.0 & 13.5 & 24.0 & 31.5 \\
Ours variant & SimpleMemVLA (Token Compression) & - & 46.0 & 13.5 & 12.5 & 18.5 & 22.6 \\
Ours variant & SimpleMemVLA (Recurrent State) & - & 31.0 & 24.0 & 17.5 & 10.0 & 20.6 \\
\midrule
\rowcolor{orange!15}
Ours & SimpleMemVLA & - & \textbf{91.5} & \textbf{96.0} & \textbf{82.5} & \textbf{83.0} & \textbf{88.3} \\
\bottomrule
\end{tabular}}
\end{table}

\textbf{Gains concentrate on tasks with greater memory demands.}
On RMBench, SimpleMemVLA is the only method in Table~\ref{tab:rmbench} whose average rises from the single-memory to the multi-memory setting, from 91.6\% to 97.0\%.
On RoboMemArena, its largest gains over the strongest overall baseline, FrameSamp+Modul, occur in Occlusion and Counting (+25.2 and +40.0 points in task success), while it does not lead on Transferring.
On MIKASA-Robo, the margin over the best prior VLA on each RememberColor task widens from 12 to 28 to 39 points as the number of candidates increases from 3 to 5 to 9.
Together, these patterns suggest that the advantage is tied to using historical evidence rather than a uniform improvement in low-level control; Section~\ref{sec:controlled} tests this attribution under a matched backbone and training setup.

\begin{table}[t]
\centering
\scriptsize
\setlength{\tabcolsep}{10pt}
\caption{\textbf{MIKASA-Robo} five-task success rates (\%). GMP is a per-task non-VLA reference. Best VLA per column in \textbf{bold}.}
\label{tab:mikasa}
\resizebox{0.99\linewidth}{!}{
\begin{tabular}{lrrrrrr}
\toprule
Method & ShellGame & Intercept & RC-3 & RC-5 & RC-9 & Avg \\
\midrule
\multicolumn{7}{l}{\emph{Vision-language-action models}} \\
CronusVLA~\citep{cronusvla} & 32 & 5 & 31 & 13 & 9 & 18.0 \\
SpatialVLA~\citep{spatialvla} & 23 & 27 & 27 & 17 & 11 & 21.0 \\
OpenVLA-OFT~\citep{openvlaoft} & 47 & 14 & 59 & 16 & 6 & 28.4 \\
$\pi_0$~\citep{pi0} & 33 & 42 & 35 & 22 & 15 & 29.4 \\
Octo~\citep{octo} & 46 & 39 & 45 & 17 & 11 & 31.6 \\
MemoryVLA~\citep{memoryvla} & 88 & 24 & 44 & 30 & 20 & 41.2 \\
MemoryVLA++~\citep{memoryvlapp} & 97 & 40 & 50 & 19 & 16 & 44.4 \\
\midrule
\multicolumn{7}{l}{\emph{Compact non-VLA memory policy (reference)}} \\
GMP~\citep{gmp} & 98 & 83 & \textbf{80} & \textbf{61} & 17 & 67.8 \\
\midrule
\rowcolor{orange!15}
SimpleMemVLA (Ours) & \textbf{99} & \textbf{83} & 71 & 58 & \textbf{59} & \textbf{74.0} \\
\bottomrule
\end{tabular}
}
\end{table}

\textbf{Preserving general-purpose manipulation performance.}
SimpleMemVLA matches the best reported average on LIBERO at 97.5\%, and the same model transfers zero-shot to LIBERO-Plus at 78.4\%, exceeding the strongest reported baseline by 5.3 points (Tables~\ref{tab:libero} and~\ref{tab:liberoplus}).
These results show that the memory gains coexist with competitive standard manipulation and strong robustness under the evaluated perturbations.

\begin{table}[t]
\centering
\scriptsize
\setlength{\tabcolsep}{5pt}
\caption{\textbf{RoboMemArena} category-level TSR and CSR (\%) over all 26 tasks under the official protocol. MemER is the benchmark authors' reimplementation, FrameSamp+Modul the only external leaderboard entry, the gray oracle row excluded from ranking. Best per column in \textbf{bold}.}
\label{tab:robomemarena}
\resizebox{0.99\linewidth}{!}{
\begin{tabular}{lcccccccccc}
\toprule
& \multicolumn{2}{c}{Transferring} & \multicolumn{2}{c}{Occlusion} & \multicolumn{2}{c}{Counting} & \multicolumn{2}{c}{Sequence} & \multicolumn{2}{c}{Average} \\
\cmidrule(lr){2-3}\cmidrule(lr){4-5}\cmidrule(lr){6-7}\cmidrule(lr){8-9}\cmidrule(lr){10-11}
Method & TSR & CSR & TSR & CSR & TSR & CSR & TSR & CSR & TSR & CSR \\
\midrule
$\pi_{0.5}$~\citep{pi05} & 20.0 & 42.8 & 12.7 & 17.2 & 14.3 & 50.9 & 60.0 & 71.6 & 21.5 & 38.7 \\
HiF-VLA~\citep{hifvla} & 17.5 & 38.9 & 12.7 & 27.1 & 8.6 & 45.9 & 42.5 & 70.2 & 16.9 & 39.8 \\
MemoryVLA~\citep{memoryvla} & 15.0 & 37.2 & 7.3 & 13.1 & 14.3 & 55.1 & 37.5 & 65.2 & 15.0 & 35.3 \\
MemER~\citep{memer} & 20.0 & 36.1 & 16.4 & 33.2 & 27.1 & 65.1 & 65.0 & 79.1 & 27.3 & 49.1 \\
FrameSamp+Modul~\citep{robomme} & \textbf{63.8} & \textbf{72.1} & 39.1 & 56.5 & 31.4 & 57.8 & 73.8 & 86.9 & 46.2 & 63.9 \\
PrediMem~\citep{robomemarena} & 22.5 & 45.2 & 27.3 & 38.4 & 45.7 & 69.3 & 72.5 & \textbf{89.5} & 38.5 & 55.2 \\
\midrule
\rowcolor{gray!15}
Ground truth (oracle)~\citep{robomemarena} & 32.5 & 54.8 & 33.6 & 49.8 & 51.4 & 75.6 & 85.0 & 92.3 & 46.1 & 64.8 \\
\midrule
\rowcolor{orange!15}
SimpleMemVLA (Ours) & 35.8 & 37.4 & \textbf{64.3} & \textbf{76.8} & \textbf{71.4} & \textbf{79.3} & \textbf{75.5} & 81.1 & \textbf{63.6} & \textbf{72.1} \\
\bottomrule
\end{tabular}
}
\end{table}

\begin{table}[t]
\centering
\scriptsize
\setlength{\tabcolsep}{14pt}
\caption{\textbf{LIBERO} success rates (\%) on the four standard suites, 500 trials per suite. Best per column in \textbf{bold}.}
\label{tab:libero}
\resizebox{0.98\linewidth}{!}{
\begin{tabular}{lccccc}
\toprule
Method & Spatial & Object & Goal & Long & Avg \\
\midrule
\multicolumn{6}{l}{\emph{General-purpose policies and VLAs}} \\
Diffusion Policy~\citep{diffusionpolicy} & 78.3 & 92.5 & 68.3 & 50.5 & 72.4 \\
TraceVLA~\citep{tracevla} & 84.6 & 85.2 & 75.1 & 54.1 & 74.8 \\
Octo~\citep{octo} & 78.9 & 85.7 & 84.6 & 51.1 & 75.1 \\
OpenVLA~\citep{openvla} & 84.7 & 88.4 & 79.2 & 53.7 & 76.5 \\
SpatialVLA~\citep{spatialvla} & 88.2 & 89.9 & 78.6 & 55.5 & 78.1 \\
$\pi_0$-FAST~\citep{pifast} & 96.4 & 96.8 & 88.6 & 60.2 & 85.5 \\
$\pi_0$~\citep{pi0} & 96.8 & 98.8 & 95.8 & 85.2 & 94.2 \\
DeepThinkVLA~\citep{deepthinkvla} & 96.6 & 99.0 & 96.4 & \textbf{96.2} & 97.0 \\
OpenVLA-OFT~\citep{openvlaoft} & 97.6 & 98.4 & 97.9 & 94.5 & 97.1 \\
RIPT-VLA~\citep{riptvla} & -- & -- & -- & -- & \textbf{97.5} \\
\midrule
\multicolumn{6}{l}{\emph{Memory-augmented VLAs}} \\
MemoryVLA~\citep{memoryvla} & \textbf{98.4} & 98.4 & 96.4 & 93.4 & 96.7 \\
CronusVLA~\citep{cronusvla} & 97.3 & \textbf{99.6} & 96.9 & 94.0 & 97.0 \\
ContextVLA~\citep{contextvla} & \textbf{98.4} & 99.0 & 97.2 & 93.4 & 97.0 \\
\midrule
\rowcolor{orange!15}
SimpleMemVLA (Ours) & 98.2 & 98.8 & \textbf{98.0} & 95.0 & \textbf{97.5} \\
\bottomrule
\end{tabular}
}
\end{table}

\begin{table}[t]
\centering
\scriptsize
\setlength{\tabcolsep}{6pt}
\caption{\textbf{LIBERO-Plus} zero-shot robustness transfer to the 10{,}030 perturbed tasks, all policies trained on standard LIBERO only. Best per column in \textbf{bold}.}
\label{tab:liberoplus}
\resizebox{0.96\linewidth}{!}{
\begin{tabular}{lcccccccc}
\toprule
Method & Camera & Robot & Lang. & Light & Backg. & Noise & Layout & Total \\
\midrule
\multicolumn{9}{l}{\emph{General-purpose policies and VLAs}} \\
OpenVLA~\citep{openvla} & 0.8 & 3.5 & 23.0 & 8.1 & 34.8 & 15.2 & 28.5 & 15.6 \\
WorldVLA~\citep{worldvla} & 0.1 & 27.9 & 41.6 & 43.7 & 17.1 & 10.9 & 38.0 & 25.0 \\
Diffusion Policy~\citep{diffusionpolicy} & 1.6 & 32.3 & 77.0 & 25.0 & 19.8 & 20.3 & 42.2 & 31.7 \\
NORA~\citep{nora} & 2.2 & 37.0 & 65.1 & 45.7 & 58.6 & 12.8 & 62.1 & 39.0 \\
UniVLA~\citep{univla} & 1.8 & 46.2 & 69.6 & 69.0 & 81.0 & 21.2 & 31.9 & 42.9 \\
Fast-WAM~\citep{fastwam} & 16.4 & 44.5 & 68.9 & 78.2 & 53.7 & 37.7 & 60.7 & 51.5 \\
$\pi_0$~\citep{pi0} & 13.8 & 6.0 & 58.8 & 85.0 & 81.4 & \textbf{79.0} & 68.9 & 53.6 \\
$\pi_0$-FAST~\citep{pifast} & 65.1 & 21.6 & 61.0 & 73.2 & 73.2 & 74.4 & 68.8 & 61.6 \\
OpenVLA-OFT$_m$~\citep{openvlaoft} & 55.6 & 21.7 & 81.0 & 92.7 & 91.0 & 78.6 & 68.7 & 67.9 \\
RIPT-VLA~\citep{riptvla} & 55.2 & 31.2 & 77.6 & 88.4 & 91.6 & 73.5 & 74.2 & 68.4 \\
OpenVLA-OFT~\citep{openvlaoft} & 56.4 & 31.9 & 79.5 & 88.7 & 93.3 & 75.8 & 74.2 & 69.6 \\
\midrule
\multicolumn{9}{l}{\emph{Memory-augmented VLAs}} \\
MemoryVLA~\citep{memoryvla} & 42.7 & 44.9 & 84.4 & 92.8 & 95.0 & 62.1 & \textbf{84.7} & 70.2 \\
MemoryVLA++~\citep{memoryvlapp} & 36.4 & \textbf{68.9} & \textbf{88.7} & 93.8 & 90.6 & 63.5 & 83.8 & 73.1 \\
\midrule
\rowcolor{orange!15}
SimpleMemVLA (Ours) & \textbf{76.9} & 68.4 & 68.6 & \textbf{96.4} & \textbf{96.8} & 74.2 & 78.0 & \textbf{78.4} \\
\bottomrule
\end{tabular}
}
\end{table}

\subsection{Real-World Evaluation}
\label{sec:real-world}

We further evaluate SimpleMemVLA on two history-dependent manipulation
tasks using a physical dual-arm robot. We fine-tune SimpleMemVLA on
180 real-robot demonstrations for Cover Blocks and 308 for Put Back
Block. At deployment, the available history grows with the episode
until it reaches a 60\,s cap, after which the policy retains the most recent 60\,s. History is sampled at 2\,fps, yielding at most 120 frames.
We deploy the policy using exact streaming inference (Section~\ref{sec:method-streaming}), which prefills the shared history while the robot executes the current action chunk, keeping decision latency manageable for closed-loop operation.

\begin{figure}[t]
\centering
\begin{minipage}[t]{\textwidth}
    \centering

    \captionof{table}{\textbf{Real-world autonomous manipulation.}
    Each initial configuration is evaluated over ten trials.
    Demonstrations denote the real-robot data used for fine-tuning.}
    \label{tab:real-world}

    \footnotesize
    \setlength{\tabcolsep}{17pt}
    \begin{tabular}{@{}lrrrr@{}}
    \toprule
    Task & Demonstrations & Configurations & Successes / Trials & SR (\%) \\
    \midrule
    Cover Blocks   & 180 & 6 & 35/60 & 58.3 \\
    Put Back Block & 308 & 4 & 28/40 & 70.0 \\
    \bottomrule
    \end{tabular}
\end{minipage}
\vfill
\begin{minipage}[t]{\textwidth}
    \centering
    \includegraphics[width=\linewidth]{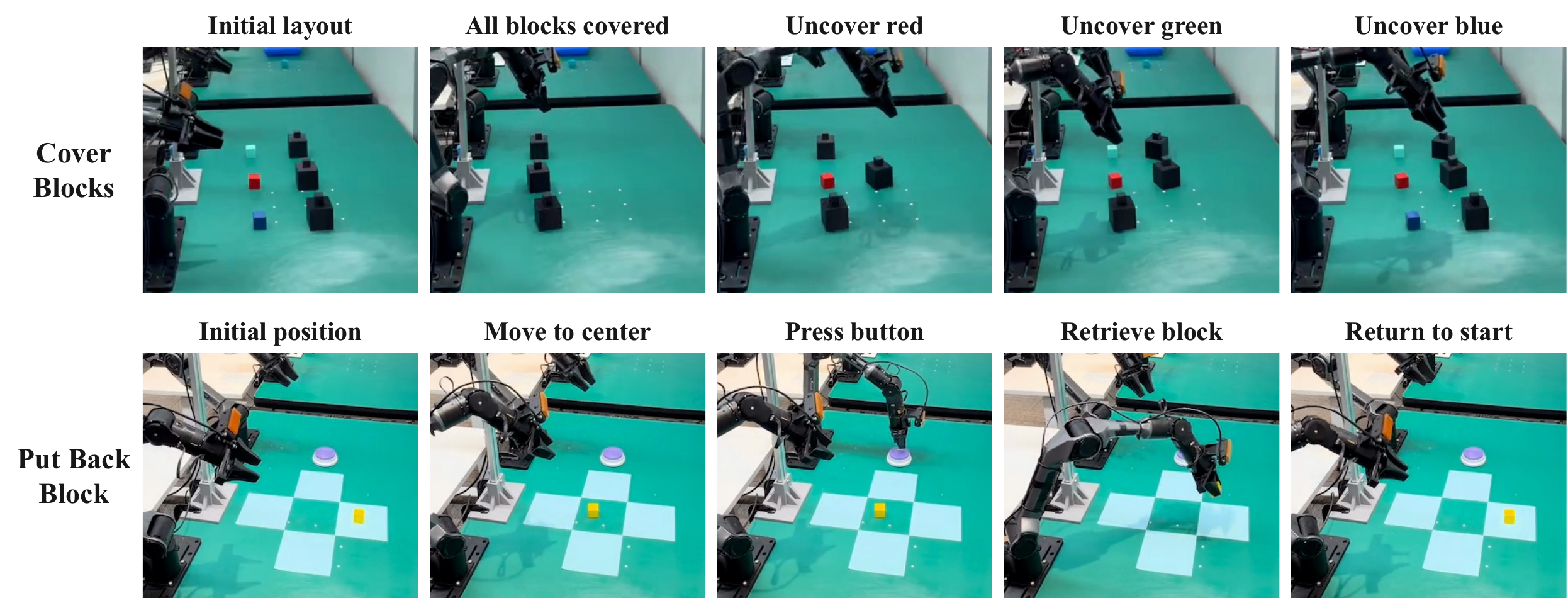}

    \caption{\textbf{Real-world autonomous rollouts.}
    Top: Cover Blocks. Bottom: Put Back Block.
    Each row shows selected frames from one successful rollout
    in chronological order.}
    \label{fig:real-world}
\end{minipage}
\end{figure}

\textbf{Tasks and results.}
We evaluate real-world versions of two RMBench tasks,
Cover Blocks and Put Back Block, which require remembering
color-to-position bindings under occlusion and an object's
initial location, respectively.
We conduct ten autonomous trials per initial configuration.
SimpleMemVLA achieves success rates of 58.3\% (35/60) across
six Cover Blocks layouts and 70.0\% (28/40) across four
Put Back Block positions (Table~\ref{tab:real-world}).
Figure~\ref{fig:real-world} presents representative autonomous
rollouts, while Appendix~\ref{app:real-world} provides
per-configuration results and implementation details.
Qualitative observations indicate that failures mainly arise from low-level execution errors, such as unsuccessful grasps.
In Cover Blocks, even after a failed grasp, the policy can still identify which covers conceal the red, green and blue blocks.
Together, these results show that native video context remains usable as memory under real-world perception and within a real-time control loop.

\subsection{Controlled Comparison of Memory Interfaces}
\label{sec:controlled}

\begin{figure}[t]
    \centering
    \includegraphics[width=\textwidth]{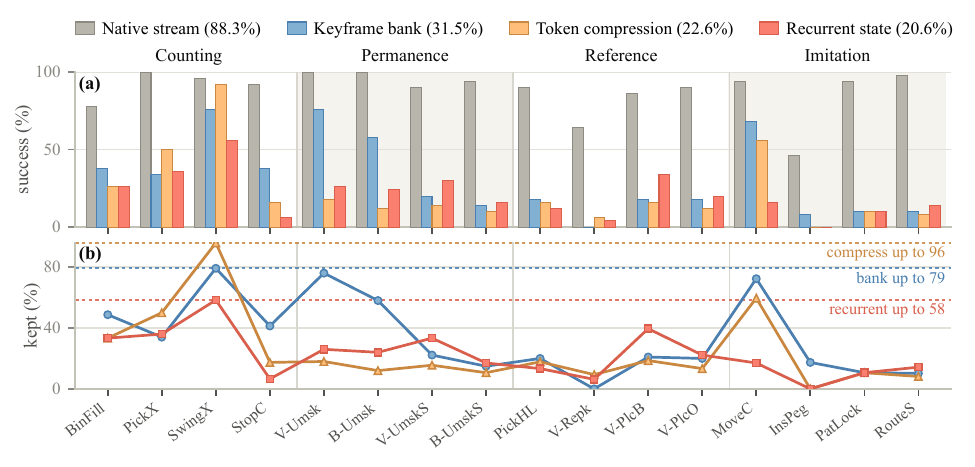}
    \caption{\textbf{Task-level effects of restricted memory interfaces on RoboMME.} The three controlled variants differ from SimpleMemVLA only in how history enters the model. 
    \emph{(a)} Success rates across 16 tasks grouped by benchmark dimension. 
    \emph{(b)} The same results normalized to native-context performance. Retrieval, token compression and recurrent state retain at most 79\%, 96\% and 58\%, respectively, with the token-compression peak confined to the count task SwingXtimes. 
    Results are from Table~\ref{tab:robomme-full}, with $n=50$ episodes per task.}
    \label{fig:variants}
\end{figure}

We compare native video context with three memory interfaces implemented on a shared stack, holding the training data, backbone, sub-task supervision, action head and optimizer fixed.
Native context reaches 88.3\% on RoboMME, compared with 31.5\% for retrieval, 22.6\% for token compression and 20.6\% for recurrent state (Table~\ref{tab:robomme}).

\textbf{Retrieval.}
Eight uniformly sampled frames preserve visual content but omit temporal order and timestamps.
The variant reaches 42.0\% on Permanence, but only 14--20\% on the two swap tasks and 10.0\% on PatternLock, where the temporal relationships between observations matter.

\textbf{Token compression.}
Compressing the full history into 64 tokens nearly matches native context on SwingXtimes (92\% versus 96\%).
However, success falls to 13.5\% on Permanence and 12.5\% on Reference, suggesting that this representation retains aggregate counts more effectively than specific past percepts.

\textbf{Recurrent state.}
A fixed 16-token state carries information forward through repeated updates, without preserving past observations for direct access.
Counting remains its strongest category at 31.0\%, while Permanence, Reference and Imitation fall to 24.0\%, 17.5\% and 10.0\%, respectively, showing that this implementation remains less effective than native context across all four categories.

These results are consistent with a cost of write-time commitment in the tested interfaces: restricting historical information before the current decision can remove evidence that the backbone subsequently needs.
Figure~\ref{fig:variants} and Appendix~\ref{app:robomme-full} provide the task-level results, with implementation details in Appendix~\ref{app:impl}.

\subsection{History Interventions}
\label{sec:analysis-read}

\begin{figure}[t]
    \centering
    \includegraphics[width=\textwidth]{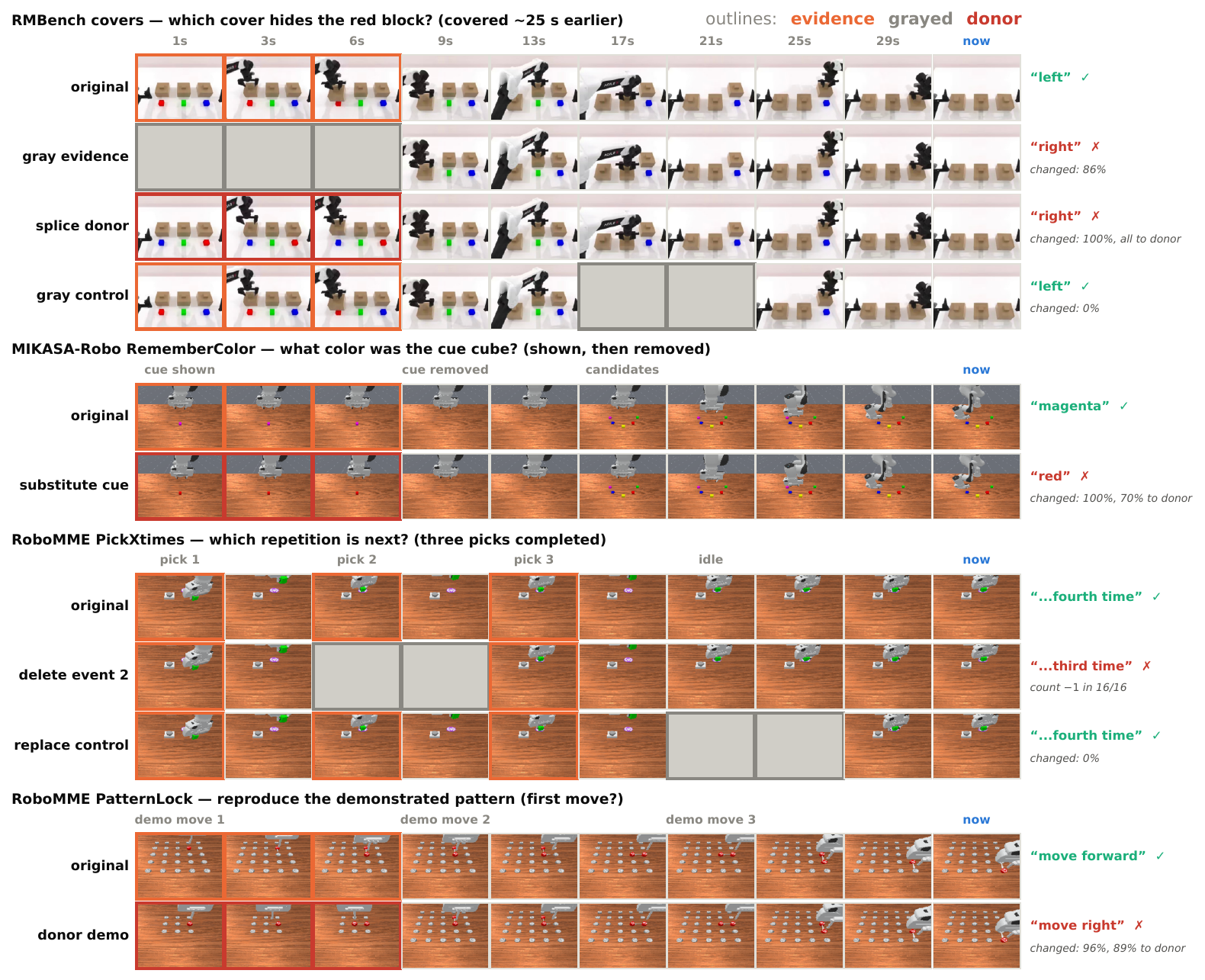}
    \caption{\textbf{History interventions redirect the output across all suites.} Blocks denote benchmarks and rows show histories from oldest to most recent, processed through the unchanged deployment pipeline. Orange borders mark evidence frames, gray fills evidence ablations and red borders donor-episode insertions. Removing evidence changes the output, donor evidence redirects it toward the donor content and matched control edits leave it unchanged.}
    \label{fig:mechanism}
\end{figure}

The benchmark results establish that SimpleMemVLA performs well on memory-dependent tasks.
We next test whether the deployed policy actually uses its visual history as memory.
At each target decision, we hold the model, instruction, current observation, robot state and deployment path fixed, modify only the historical video, and regenerate the policy output.

\textbf{Removing task-relevant evidence.}
Masking the history affects the policy specifically when the removed frames contain evidence required by the current decision.
On the RMBench cover-blocks task, masking the historical video segment that records where the red block was covered causes the policy to select the wrong cover, whereas masking task-irrelevant history leaves its output unchanged (Figure~\ref{fig:mechanism} row 2 and 4).
On MIKASA-Robo RememberColor, masking the historical frames that contain the color cue causes the policy to select the wrong color, whereas masking task-irrelevant frames leaves its output unchanged. 
Together with the RMBench result, this shows that the policy fails specifically when task-relevant visual memory is removed, rather than in response to visual masking itself.

RoboMME PickXtimes provides a more fine-grained test. 
Masking one completed pick-and-place event causes the policy to infer that one fewer repetition has occurred---for example, changing its output from ``the fourth time'' to ``the third time''---and therefore to execute one additional pick. 
Masking a matched history segment that contains no pick-and-place event leaves the inferred count unchanged. 
The policy therefore remembers the exact number of completed events, rather than merely whether a relevant event has occurred (Figure~\ref{fig:mechanism} row 8 and 9).

\textbf{Replacing historical evidence.}
On RMBench, replacing the original covering event with frames from another rollout in which the red block is placed under a different cover causes the policy to select the cover shown in the replacement video. 
Replacing it with frames showing the same cover leaves the decision unchanged (Figure~\ref{fig:mechanism} row 3).
On MIKASA-Robo, replacing a magenta cue with a red cue causes the policy to grasp the red block (Figure~\ref{fig:mechanism} row 6).
On RoboMME PatternLock, replacing a demonstration beginning with ``move forward'' by one beginning with ``move right'' causes the policy to execute ``move right,'' whereas replacing it with another demonstration of the same first move leaves the behavior unchanged (Figure~\ref{fig:mechanism} row 11).

Neither type of edited history appears during training.
The replacement conditions splice a visual segment from one rollout into the remaining history of another, while the PickXtimes condition masks one completed event in an otherwise valid rollout.
Without intervention-specific supervision or parameter updates, the frozen policy nevertheless interprets these newly constructed visual histories at inference time: it follows the substituted location, color, and demonstration, and adjusts its count to the completed events that remain visible.

This behavior reveals an emergent visual in-context learning capability. 
The frozen policy can infer the task state or intended behavior directly from a newly constructed visual context and adapt its current action accordingly. 
Together, the masking and replacement experiments show that native context serves both as genuine memory of past events and as an inference-time visual interface through which new evidence can re-specify the policy's behavior.

\begin{figure}[t]
    \centering
    \includegraphics[width=\textwidth]{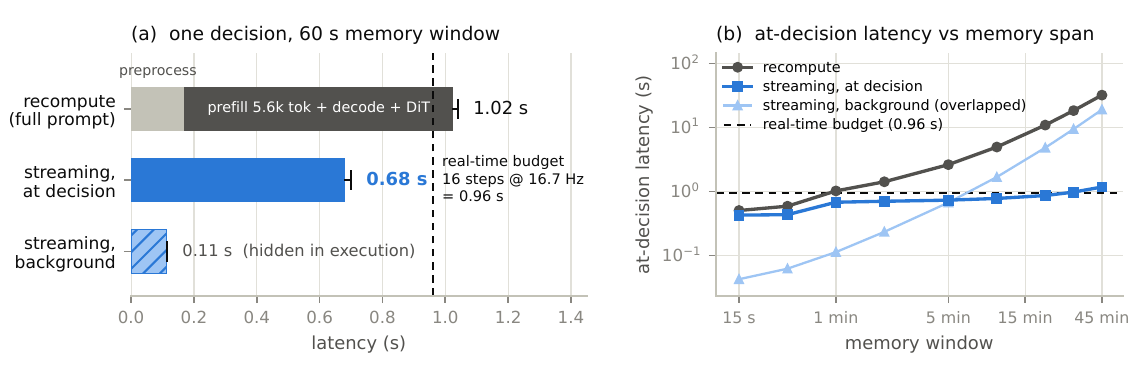}
    \caption{\textbf{Streaming inference reduces decision latency.}
    Measured on one H100 (bf16, batch size 1).
    \emph{(a)} With 60\,s histories, streaming reduces decision latency from 1.02\,s to 0.68\,s, close to the same-model single-frame baseline of approximately 0.65\,s.
    \emph{(b)} On artificially constructed 45\,min inputs (245k tokens), decision-path latency falls from 32.1\,s to 1.18\,s.}
    \label{fig:latency}
\end{figure}

\subsection{Streaming Inference with Prefix Prefill}
\label{sec:streaming}

The preceding results establish both the effectiveness of SimpleMemVLA and its use of visual history.
The remaining practical question is the cost of processing that history.
With a 60\,s context window, full recomputation processes approximately 5.6k tokens at every decision, compared with approximately 0.5k tokens for single-frame input.

The streaming implementation of Section~\ref{sec:method-streaming} uses a shared-prefix cache to reduce decision latency while producing the same generated sub-task (Figure~\ref{fig:arch}b).
On one H100, for the representative minute-scale configuration evaluated in Figure~\ref{fig:latency}a, streaming reduces decision latency from 1.02\,s to 0.68\,s, close to the approximately 0.65\,s latency of the same model with single-frame input.

In our scaling experiment, an artificially constructed input corresponding to 45\,min of history contains approximately 245k tokens, near the backbone's 262k-token context limit.
At this input length, full-recomputation latency reaches 32.1\,s per decision, whereas the streamed decision path requires only 1.18\,s (Figure~\ref{fig:latency}b).
The key--value cache size depends on the retained context length and remains bounded under the standard 60\,s history cap.
Appendix~\ref{app:swa} presents a sliding-window-attention (SWA) variant that uses the same bounded attention window during training and inference.
At deployment, it appends one new temporal patch and evicts one expired cache block at every decision, keeping cache state independent of the total episode length (Figure~\ref{fig:swa-method}).

\begin{figure}[t]
    \centering
    \includegraphics[width=\textwidth]{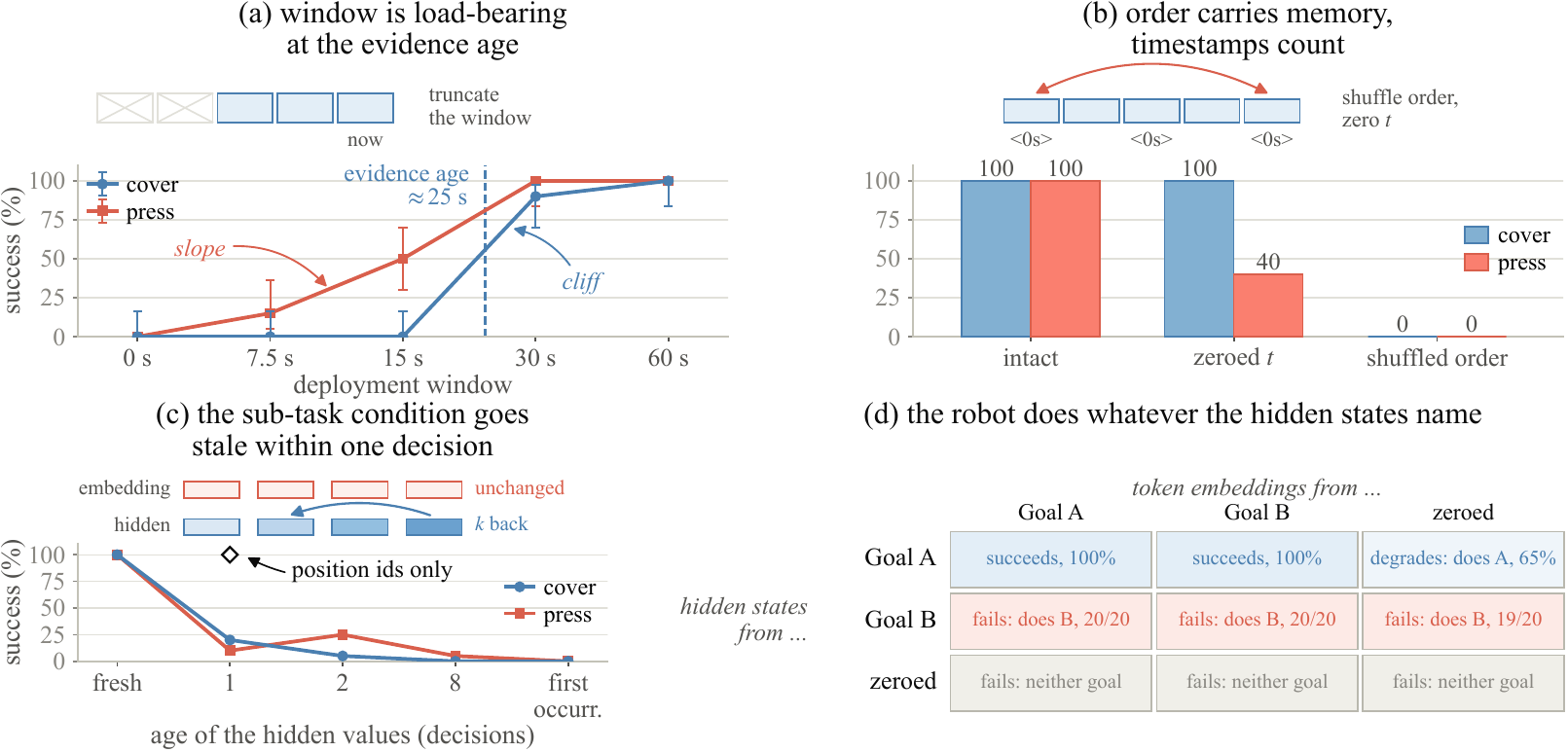}
    \caption{\textbf{Native-context memory relies on retained temporal evidence and contextual hidden states.} \emph{(a)} Cover-blocks fails once the covering event leaves the window, whereas press-button degrades gradually as completed presses are removed. \emph{(b)} Both tasks depend on frame order, while timestamps matter primarily for counting. \emph{(c)} Stale sub-task hidden states sharply reduce success, whereas equally stale position IDs leave it unchanged. \emph{(d)} Behavior follows the source of the hidden states rather than the token embeddings, identifying contextual hidden states as the memory-to-action interface.}
    \label{fig:ablations}
\end{figure}

\subsection{Ablations: Dissecting the Memory Pathway}
\label{sec:ablations}

We conduct the ablations on two RMBench tasks that represent complementary memory requirements: cover-blocks requires recalling a single past event, whereas press-button requires counting repeated events throughout the trajectory (Figure~\ref{fig:ablations}).
\textbf{The evidence must remain in the window.}
On cover-blocks, a 30\,s window retains the covering event from approximately 25\,s earlier, whereas a 15\,s window excludes it and causes failure.
On press-button, shorter windows progressively remove completed presses and reduce success.
Both tasks fail with only the current frame (Figure~\ref{fig:ablations}a).
\textbf{Order and timestamps provide complementary temporal information.}
Shuffling frame order causes both tasks to fail.
Setting the plaintext timestamps to zero leaves cover-blocks unaffected but substantially degrades press-button, indicating that explicit timing provides additional information for counting repeated events (Figure~\ref{fig:ablations}b).
\textbf{Hidden states carry the history-dependent control signal.}
Replacing token embeddings with those of another target largely preserves the selected behavior, whereas substituting hidden states redirects it toward the substituted target; zeroing the hidden states causes failure.
Contextual hidden states therefore carry the primary history-dependent control signal, while token embeddings support stable execution (Figure~\ref{fig:ablations}d).
\textbf{The representation must be refreshed for each decision.}
Deleting the target word makes the visible sub-task identical across decisions, yet the policy still follows the required cover sequence and press count.
Reusing hidden states from even one earlier decision sharply reduces success, whereas equally stale position IDs do not (Figure~\ref{fig:ablations}c).
The backbone forms this representation by reading the retained history after the current decision context is available, selecting evidence at read time rather than deciding in advance what to preserve for future decisions.

\section{Conclusion}
\label{sec:conclusion}
We have shown that a pretrained VLM's native video context can serve as the memory of a VLA without a dedicated memory module. Because the history is kept intact until each decision, the policy avoids write-time commitment, and this alone yields state-of-the-art results on four memory benchmarks, a wide margin over retrieval, compression and recurrent-state mechanisms under the same backbone and training setup, and autonomous history-dependent manipulation on a physical dual-arm robot. Prefilling the shared history during action execution keeps decision latency close to that of a single-frame VLA.
Although our findings are limited to the evaluated backbones, benchmarks and memory interfaces, we argue that native context should serve as a matched baseline for future VLA memory mechanisms.
Before adding dedicated memory machinery, the first comparison should be the same policy given access to its own visual history.

% \subsection*{AI use statement}

% (This section is \textbf{required} and does not count toward the page limit.)

% Generative AI tools were used solely for language polishing. All AI-assisted edits were reviewed and approved by the authors, who take full responsibility for the final content of this work.

% \subsection*{Ethics statement}

% This work develops and evaluates robot manipulation policies exclusively in simulation benchmarks. It involves no human subjects or personal data. We identify no specific ethical or dual-use concerns beyond those generally associated with robot learning research.

% \subsection*{Reproducibility statement}

% To support reproducibility, our code is publicly available at \url{https://github.com/wadeKeith/SimpleMemVLA}. We also provide the training configurations and evaluation protocols used in this work. Section~\ref{sec:method} describes the complete input construction procedure, including sampling rates, timestamps, and modality rules, as well as the implementation shared between training and inference. Section~\ref{sec:experiments} details the benchmarks, evaluation seeds, and protocols. All reported evaluations use held-out seeds.

% \subsubsection*{Author Contributions}
% \tofill{Optional; fill at camera-ready.}

% \subsubsection*{Acknowledgments}
% \tofill{Fill at camera-ready; funding acknowledgements go here.}

% \newpage
\bibliography{iclr2027_conference}
\bibliographystyle{iclr2027_conference}

\appendix
\section{Implementation Details}
\label{app:impl}

Table~\ref{tab:impl} lists the per-benchmark instantiation of the configuration tuple from Section~\ref{sec:method-streaming}.
All suites share the same backbone (Qwen3.5-4B, native video interface with native plaintext timestamps), the same conditioning rule and the same losses.
Optimization uses AdamW with separate learning rates for the backbone and the action head ($10^{-5}$ and $5\times10^{-5}$), cosine decay and gradient clipping.
The stride $s$ is the integer rounding of $f_c/f_v$.

\begin{table}[t]
\caption{Per-benchmark instantiation of the SimpleMemVLA configuration tuple and the resulting training cost. Everything else about the method is identical across suites. Training time is wall-clock hours on 128 H100 GPUs.}
\label{tab:impl}
\begin{center}
\footnotesize
\setlength{\tabcolsep}{4pt}
\resizebox{\textwidth}{!}{%
\begin{tabular}{lccccc}
\toprule
 & RMBench & RoboMME & MIKASA-Robo & RoboMemArena & LIBERO \\
\midrule
Embodiment & Aloha-AgileX, bimanual & Panda, single & Panda, single & Franka, single & Franka, single \\
Action dimension $d_a$ & 14 & 8 & 8 & 7 & 7 \\
History camera $\mathcal{C}_{\mathrm{hist}}$ & head & front & overhead & front & agentview \\
Current cameras $\mathcal{C}_{\mathrm{cur}}$ & both wrists & wrist & wrist & wrist & wrist \\
Window $T_w$ (s) & 60 & 60 & 3 & 126 & 30 \\
Video rate $f_v$ (fps) & 2 & 2 & 20 & 1 & 2 \\
Frames $K$ & 120 & 120 & 60 & 126 & 60 \\
Stride $s$ & 8 & 10 & 1 & 20 & 10 \\
Action horizon $H$ & 30 & 30 & 16 & 16 & 16 \\
\midrule
Training time (h) & 23 & 20 & 16 & 28 & 14 \\
\bottomrule
\end{tabular}}
\end{center}
\end{table}

\paragraph{Recurrent-state training.}
The recurrent variant maintains a 16-token state, with gradients propagated through unrolls of four consecutive decisions.
Finite-horizon backpropagation limits training cost~\citep{williams1990tbptt}; recurrent memory transformers likewise treat the unroll length as a training hyperparameter~\citep{bulatov2022rmt}.
Our results characterize the evaluated implementation under this training protocol.

\paragraph{Sub-task labels.}
The sub-task label $g^*_t$ of Section~\ref{sec:method-head} is written offline by a cloud VLM: shown each demonstration, it states a one-sentence sub-task for every supervision anchor and the dataset stores the result per frame.
The three mechanism variants of Section~\ref{sec:controlled} are trained on these identical labels and the deployed policy never queries the annotator, so annotation is a training-data cost shared by every within-stack comparison and absent at inference.

\paragraph{Benchmarks and metrics.}
\textbf{RMBench}~\citep{rmbench} comprises ten memory-dependent bimanual tasks built on RoboTwin~2.0~\citep{robotwin}, grouped into single-memory $M(1)$ and multi-memory $M(n)$ settings.
Cross-method averages exclude Place-Mat, for which no prior baseline is reported, and cover the remaining nine tasks.
Most published RMBench baselines train a separate specialist for each task.
\textbf{RoboMME}~\citep{robomme} contains sixteen single-arm tasks spanning Counting, Permanence, Reference and Imitation, and reports fourteen memory variants built on a shared $\pi_{0.5}$ backbone.
\textbf{MIKASA-Robo}~\citep{mikasa} comprises five single-arm tasks testing memory under occlusion and partial observability.
\textbf{RoboMemArena}~\citep{robomemarena} contains twenty-six single-arm tasks across Transferring, Occlusion, Counting and Sequence, with trajectories averaging more than one thousand control steps.
It reports all-stage task success rate (TSR) and stage completion rate (CSR).

\textbf{LIBERO}~\citep{libero} evaluates general-purpose manipulation across the Spatial, Object, Goal and Long suites.
\textbf{LIBERO-Plus}~\citep{liberoplus} expands these suites into 10{,}030 perturbed tasks across seven dimensions.
We evaluate the LIBERO-trained model without further training; the detailed transfer protocol and breakdowns are provided in Appendix~\ref{app:liberoplus-detail}.

\paragraph{Baseline provenance and per-suite protocols.}
RMBench third-party rows come from the papers cited in the text and EventVLA is quoted in its visual-anchors-only configuration. RoboMME rows follow the benchmark report, except the third-party WeaveLA on its own retrained $\pi_{0.5}$ stack. RoboMemArena rows follow the benchmark report, with the external FrameSamp+Modul entry quoted from the official leaderboard and evaluation run under the official protocol of 51 rollouts per task. MIKASA-Robo and LIBERO-Plus VLA baselines follow the MemoryVLA++ report~\citep{memoryvlapp} and GMP numbers come from its Figure~11 with per-task training and tuned horizons. LIBERO numbers for Diffusion Policy, Octo, OpenVLA, $\pi_0$ and $\pi_0$-FAST follow the OpenVLA-OFT report~\citep{openvlaoft} at 500 trials per suite and RIPT-VLA is quoted at its reported four-suite average. RMBench uses 100 evaluation seeds per task. Instructions are frozen to each benchmark's own task strings throughout, so no prompt is tuned per method.

Two engineering bounds delimit how far the native window stretches. The backbone's 262k position limit corresponds to about 48\,min of history at the deployment sampling rate. The vision processor further enforces a total pixel budget that would shrink per-frame resolution once the window exceeds a few minutes, which we lift at runtime by scaling its longest-edge budget with the frame count so that every frame keeps the production resolution.

\section{Real-World Evaluation Details}
\label{app:real-world}

\paragraph{Robot and observations.}
We use a dual-arm Piper robot equipped with one head-mounted
camera and two wrist-mounted cameras.
Following the input construction in Section~\ref{sec:method},
head-camera observations provide the visual history,
while the two wrist cameras provide current observations.

\paragraph{Tasks and evaluation.}
In Cover Blocks, the robot first covers all three colored
blocks with identical lids and then uncovers them in
red--green--blue order.
In Put Back Block, the robot moves the block from its initial
mat to the center, presses the button, retrieves the block
and returns it to the same mat.
We evaluate all six color arrangements for Cover Blocks
and four initial positions for Put Back Block,
with ten autonomous trials per configuration.

\paragraph{Fine-tuning data.}
Fine-tuning uses 180 real-robot demonstrations for Cover Blocks
and 308 for Put Back Block.

\paragraph{Deployment.}
Following the RMBench action representation, the policy
predicts 14-dimensional actions comprising absolute
joint-position targets and one gripper command per arm.
Each prediction produces a chunk of $30$ actions, of which
the first $16$ are executed.
Actions are executed at 25\,Hz, matching the sampling rate
of the demonstration data.
The head-camera history grows with the episode up to
a 60\,s cap, after which it retains the most recent 60\,s.
History is sampled at 2\,fps, yielding at most 120 frames,
and processed using exact streaming inference.

\begin{table}[t]
\centering
\small
\caption{\textbf{Real-world results by initial configuration.}
Each entry reports successes over ten autonomous trials.
Cover Blocks labels list colors from far to near.
Put Back Block labels follow the task coordinate convention.}
\label{tab:real-world-configs}
\setlength{\tabcolsep}{14pt}
\begin{tabular}{@{}lrrrrrrr@{}}
\toprule
Cover Blocks & RGB & GRB & GBR & RBG & BRG & BGR & Overall \\
\midrule
Successes/trials & 8/10 & 6/10 & 7/10 & 6/10 & 5/10 & 3/10 & 35/60 \\
\bottomrule
\end{tabular}

\vspace{6pt}
\setlength{\tabcolsep}{23.5pt}
\begin{tabular}{@{}lrrrrr@{}}
\toprule
Put Back Block & Up & Down & Left & Right & Overall \\
\midrule
Successes/trials & 7/10 & 9/10 & 4/10 & 8/10 & 28/40 \\
\bottomrule
\end{tabular}
\end{table}

\begin{figure}[t]
\centering
\includegraphics[width=\textwidth]{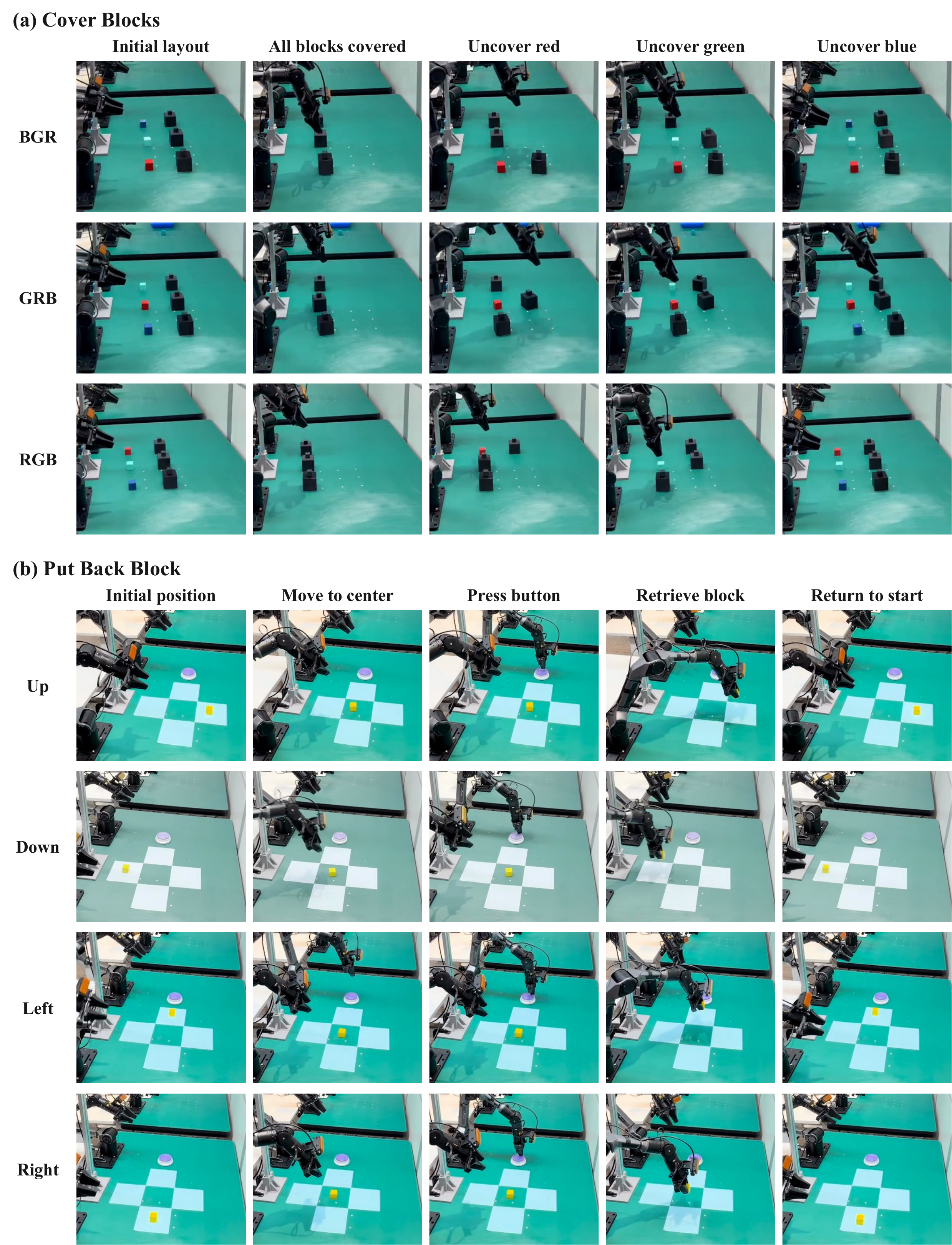}
\caption{\textbf{Additional real-world autonomous rollouts.}
(a) Cover Blocks under three initial color arrangements,
labeled from far to near.
(b) Put Back Block from four initial positions, labeled using
the task coordinate convention rather than camera-image directions.
Each row shows chronological frames from one successful rollout.}
\label{fig:real-world-appendix}
\end{figure}

Figure~\ref{fig:real-world-appendix} shows additional successful
autonomous rollouts. These selected examples are separate from
the aggregate evaluation in Table~\ref{tab:real-world-configs}.

\section{Full RoboMME Results}
\label{app:robomme-full}

Table~\ref{tab:robomme-full} reports the complete task-level RoboMME results underlying Table~\ref{tab:robomme}, including the per-task scores of our controlled variants, and Figure~\ref{fig:variants} plots the three rebuilds against the native stream. Figures~\ref{fig:robomme-representative} and~\ref{fig:robomme-all-methods} visualize the benchmark pool.

\begin{figure}[t]
\centering
\includegraphics[width=\textwidth]{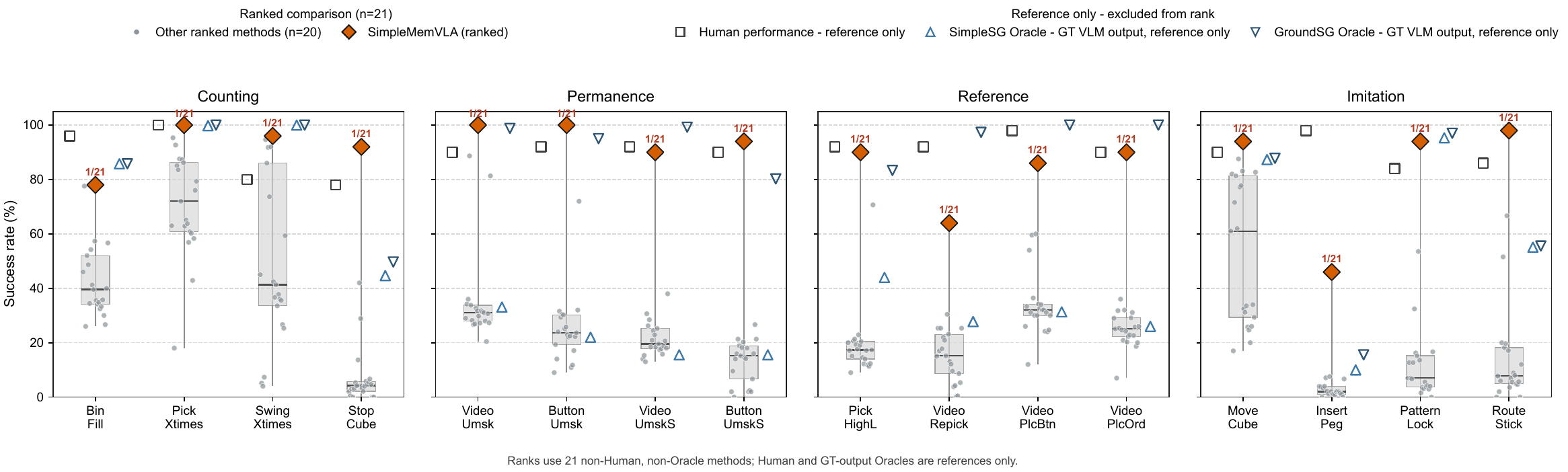}
\caption{\textbf{First on all sixteen RoboMME tasks among the 21 deployable methods.} Gray dots denote the 20 other non-Human, non-Oracle methods evaluated by the benchmark. Boxes show the interquartile range of the 21-method ranked pool, with median and full range. Orange diamonds mark SimpleMemVLA and labels report its competition rank out of 21. Human performance and the two symbolic Oracle variants, which replace VLM outputs with ground-truth outputs, are shown only as references and are excluded from all ranks and distribution statistics.}
\label{fig:robomme-position}
\end{figure}

\begin{figure}[t]
\centering
\includegraphics[width=\textwidth]{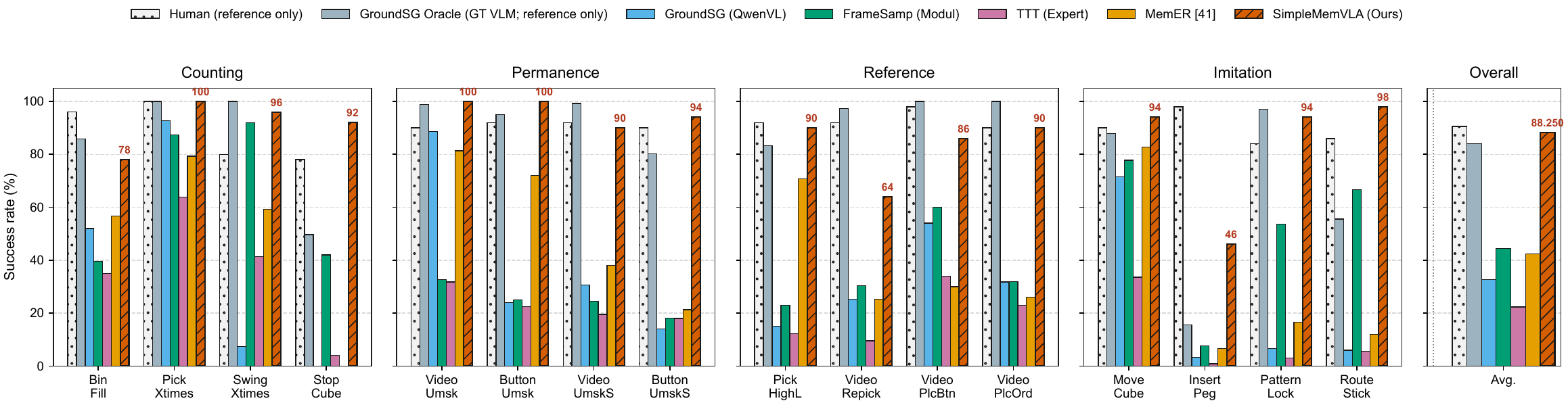}
\caption{Task-wise RoboMME success rates (\%, higher is better) for representative methods. Panels group the 16 tasks into Counting, Permanence, Reference and Imitation. Overall is recomputed from all 16 task scores. Human and GroundSG (Oracle, using ground-truth VLM outputs) are shown only as references and are excluded from the ranked comparison. For each remaining baseline family, the row with the highest overall AVG is shown. Values above orange hatched bars report SimpleMemVLA scores.}
\label{fig:robomme-representative}
\end{figure}

\begin{figure}[p]
\centering
\includegraphics[width=0.96\textwidth]{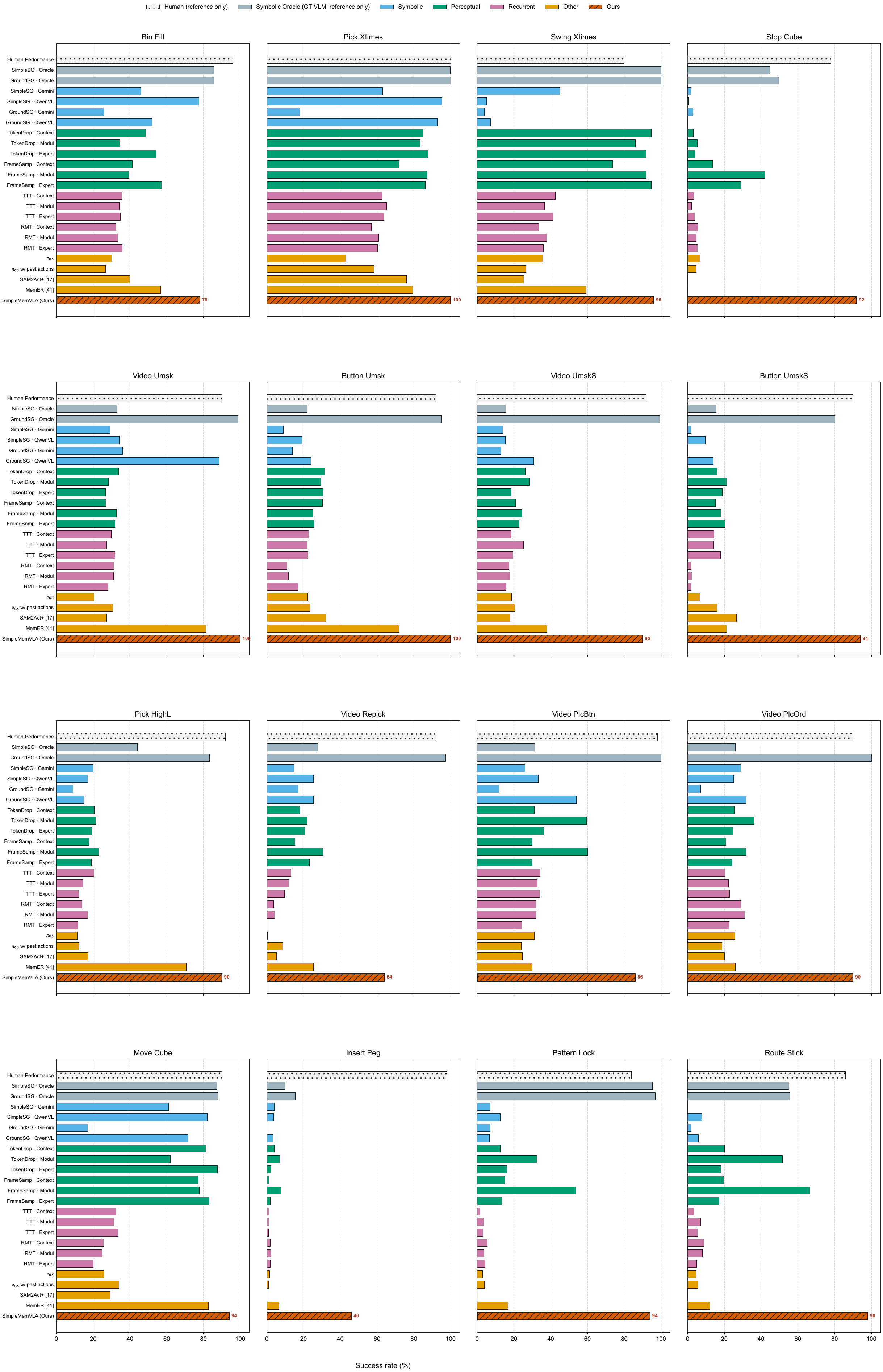}
\caption{Complete task-wise RoboMME success rates (\%, higher is better) for all 24 rows. Tasks follow the source-table order and methods retain the same order in every panel. Colors encode memory families. The orange hatched bar denotes SimpleMemVLA.}
\label{fig:robomme-all-methods}
\end{figure}

\begin{landscape}
\begin{table}[p]
\caption{Complete task-level RoboMME success rates (\%). Human and GT-VLM Oracle rows (gray) are reference-only and excluded from ranking. \textbf{Bold} and \underline{underline} mark ranks 1 and 2 among the remaining 21 methods. Task abbreviations follow Table~\ref{tab:robomme}'s four dimensions in source order. The three SimpleMemVLA variant rows are the controlled within-stack rebuilds of Section~\ref{sec:controlled} and are excluded from ranking.}
\label{tab:robomme-full}
\begin{center}
\resizebox{\linewidth}{!}{%
\begin{tabular}{lllrrrrrrrrrrrrrrrrr}
\toprule
Memory & Method & Integr.\ / VLM & \multicolumn{4}{c}{Counting} & \multicolumn{4}{c}{Permanence} & \multicolumn{4}{c}{Reference} & \multicolumn{4}{c}{Imitation} & AVG \\
\cmidrule(lr){4-7}\cmidrule(lr){8-11}\cmidrule(lr){12-15}\cmidrule(lr){16-19}
 &  &  & BinFill & PickX & SwingX & StopC & V-Umsk & B-Umsk & V-UmskS & B-UmskS & PickHL & V-Repk & V-PlcB & V-PlcO & MoveC & InsPeg & PatLock & RouteS & \\
\midrule
\rowcolor{gray!15}
Human & Human Performance & - & 96 & 100 & 80 & 78 & 90 & 92 & 92 & 90 & 92 & 92 & 98 & 90 & 90 & 98 & 84 & 86 & 90.5 \\
\midrule
\rowcolor{gray!15}
Symbolic (Oracle) & SimpleSG~\citep{robomme} & GT VLM & 85.78 & 99.78 & 100 & 44.67 & 33.11 & 22 & 15.56 & 15.56 & 44 & 27.78 & 31.33 & 26 & 87.33 & 10 & 95.33 & 55.11 & 49.58 \\
\rowcolor{gray!15}
Symbolic (Oracle) & GroundSG~\citep{robomme} & GT VLM & 85.78 & 100 & 100 & 49.67 & 98.78 & 95 & 99.22 & 80.22 & 83.33 & 97.33 & 100 & 100 & 87.78 & 15.56 & 97 & 55.56 & 84.08 \\
\midrule
Symbolic & SimpleSG & Gemini & 46 & 63 & 45 & 2 & 29 & 9 & 14 & 2 & 20 & 15 & 26 & 29 & 61 & 4 & 7 & 0 & 23.25 \\
Symbolic & SimpleSG & QwenVL & \underline{77.56} & \underline{95.33} & 5.11 & 0.44 & 34.22 & 19.33 & 15.33 & 9.56 & 17.11 & 25.33 & 33.33 & 25.11 & 82 & 3.78 & 12.67 & 7.78 & 29 \\
Symbolic & GroundSG & Gemini & 26 & 18 & 4 & 3 & 36 & 14 & 13 & 0 & 9 & 17 & 12 & 7 & 17 & 0 & 7 & 2 & 11.56 \\
Symbolic & GroundSG & QwenVL & 52 & 92.67 & 7.33 & 0 & \underline{88.67} & 24 & 30.67 & 14 & 15.11 & 25.33 & 54 & 31.78 & 71.56 & 3.33 & 6.67 & 6 & 32.7 \\
\midrule
Perceptual & TokenDrop~\citep{robomme} & Context & 48.67 & 85.11 & \underline{94.67} & 3.11 & 33.78 & 31.56 & 26.22 & 16 & 20.67 & 17.78 & 31.11 & 25.33 & 81.33 & 4 & 12.67 & 20 & 34.5 \\
Perceptual & TokenDrop & Modul & 34.44 & 83.56 & 86 & 5.33 & 28.22 & 29.33 & 28.44 & 21.33 & 21.33 & 22 & 59.56 & \underline{36} & 62 & 7.11 & 32.44 & 51.56 & 38.04 \\
Perceptual & TokenDrop & Expert & 54.22 & 87.56 & 91.78 & 4.22 & 26.67 & 30.44 & 18.44 & 18.89 & 19.33 & 20.89 & 36.44 & 24.67 & \underline{87.56} & 2.22 & 16.22 & 18.22 & 34.86 \\
Perceptual & FrameSamp~\citep{robomme} & Context & 41.22 & 72 & 73.67 & 13.67 & 26.89 & 30.22 & 20.89 & 15.22 & 17.67 & 15.22 & 30 & 20.89 & 77.22 & 1.22 & 15.22 & 19.67 & 30.68 \\
Perceptual & FrameSamp & Modul & 39.56 & 87.33 & 92 & \underline{42} & 32.67 & 25.11 & 24.44 & 18.22 & 22.89 & \underline{30.44} & \underline{60} & 32 & 77.78 & \underline{7.56} & \underline{53.56} & \underline{66.67} & \underline{44.51} \\
Perceptual & FrameSamp & Expert & 57.33 & 86.22 & \underline{94.67} & 28.89 & 31.78 & 25.78 & 22.89 & 20.22 & 19.11 & 23.11 & 30 & 24.22 & 83.11 & 2 & 13.56 & 17.11 & 36.25 \\
\midrule
Recurrent & TTT~\citep{robomme} & Context & 35.56 & 62.89 & 42.44 & 3.33 & 29.78 & 22.89 & 18.44 & 14.44 & 20.44 & 13.11 & 34.22 & 20.22 & 32.44 & 1.11 & 1.56 & 3.56 & 22.28 \\
Recurrent & TTT & Modul & 34.22 & 65.11 & 36.67 & 2.11 & 27.22 & 22.11 & 25.22 & 14.11 & 14.56 & 12.11 & 32.67 & 22.33 & 31.22 & 1.11 & 3.56 & 7 & 21.96 \\
Recurrent & TTT & Expert & 34.89 & 63.78 & 41.33 & 4 & 31.78 & 22.44 & 19.56 & 18 & 12.22 & 9.56 & 34 & 22.89 & 33.56 & 0.89 & 3.11 & 5.56 & 22.35 \\
Recurrent & RMT~\citep{robomme} & Context & 32.44 & 56.89 & 33.56 & 5.78 & 31.33 & 10.89 & 17.33 & 2 & 14 & 3.78 & 32 & 29.11 & 25.78 & 2 & 5.56 & 8.89 & 19.46 \\
Recurrent & RMT & Modul & 33.33 & 60.78 & 37.78 & 4.67 & 31.11 & 11.78 & 17.78 & 2.44 & 17.11 & 4.22 & 32 & 31.11 & 24.67 & 2.21 & 3.78 & 8 & 20.17 \\
Recurrent & RMT & Expert & 35.78 & 60.22 & 36 & 5.56 & 28 & 17.11 & 15.78 & 2 & 11.78 & 0.22 & 24.22 & 22.67 & 20 & 1.89 & 4.22 & 4.89 & 18.15 \\
\midrule
Other & $\pi_{0.5}$~\citep{pi05} & - & 30 & 42.89 & 35.56 & 6.67 & 20.44 & 22.22 & 18.67 & 6.67 & 11.33 & 0.44 & 31.11 & 25.78 & 26 & 1.56 & 2.89 & 4.67 & 17.93 \\
Other & $\pi_{0.5}$ w/ past actions~\citep{robomme} & - & 26.67 & 58.33 & 26.67 & 4.67 & 30.67 & 23.67 & 20.67 & 16 & 12.33 & 8.67 & 24 & 18.67 & 34 & 1 & 4 & 5.67 & 19.73 \\
Other & SAM2Act+~\citep{sam2act} & - & 40 & 76 & 25.33 & 0 & 27.33 & 32 & 18 & \underline{26.67} & 17.33 & 5.33 & 24.67 & 20 & 29.33 & 0 & 0 & 0 & 21.37 \\
Other & MemER~\citep{memer} & - & 56.67 & 79.33 & 59.33 & 0 & 81.33 & \underline{72} & \underline{38} & 21.33 & \underline{70.67} & 25.33 & 30 & 26 & 82.67 & 6.67 & 16.67 & 12 & 42.38 \\
\midrule
Ours variant & SimpleMemVLA (Retrieval) & - & 38 & 34 & 76 & 38 & 76 & 58 & 20 & 14 & 18 & 0 & 18 & 18 & 68 & 8 & 10 & 10 & 31.5 \\
Ours variant & SimpleMemVLA (Token Compression) & - & 26 & 50 & 92 & 16 & 18 & 12 & 14 & 10 & 16 & 6 & 16 & 12 & 56 & 0 & 10 & 8 & 22.625 \\
Ours variant & SimpleMemVLA (Recurrent State) & - & 26 & 36 & 56 & 6 & 26 & 24 & 30 & 16 & 12 & 4 & 34 & 20 & 16 & 0 & 10 & 14 & 20.625 \\
\midrule
\rowcolor{orange!15}
Ours & SimpleMemVLA & - & \textbf{78} & \textbf{100} & \textbf{96} & \textbf{92} & \textbf{100} & \textbf{100} & \textbf{90} & \textbf{94} & \textbf{90} & \textbf{64} & \textbf{86} & \textbf{90} & \textbf{94} & \textbf{46} & \textbf{94} & \textbf{98} & \textbf{88.25} \\
\bottomrule
\end{tabular}}
\end{center}
\end{table}
\end{landscape}

\section{LIBERO-Plus: Protocol Details and Breakdowns}
\label{app:liberoplus-detail}

\paragraph{Protocol.}
We follow the official LIBERO-Plus protocol~\citep{liberoplus}: one trial per task over all 10{,}030 tasks (zero exclusions), a frozen initial state and fixed environment seed per task, ten no-op settling steps, per-suite step budgets of 220/280/300/520 and success judged by the simulator's built-in check.
The seven dimensions differ in task counts (Camera 1{,}599, Robot 1{,}550, Language 1{,}537, Light 1{,}142, Background 1{,}076, Noise 1{,}601, Layout 1{,}525), so the Total in Table~\ref{tab:liberoplus} is the success rate pooled over all tasks rather than a mean of the seven columns.
One detail matters when reading the Language column: for the six non-language dimensions the official instruction string appends the perturbation tag to the task instruction (e.g., ``\ldots\ table 1''), which we feed verbatim.
Only the Language dimension replaces the instruction with a rewritten one, so that column specifically measures robustness to paraphrase.

\begin{table}[t]
\caption{\textbf{LIBERO-Plus breakdowns.} Top: by source LIBERO suite. Bottom: by the benchmark's five difficulty levels, which stratify tasks by the accuracy of four reference models~\citep{liberoplus}. The 121 tasks that carry no difficulty label in the benchmark metadata are omitted from the bottom panel.}
\label{tab:liberoplus-breakdown}
\begin{center}
\small
\begin{tabular}{lccccc}
\toprule
\multicolumn{6}{l}{\emph{By source suite}} \\
 & Spatial & Object & Goal & Long & \\
Tasks & 2{,}402 & 2{,}518 & 2{,}591 & 2{,}519 & \\
Success (\%) & 81.8 & 83.9 & 75.0 & 73.1 & \\
\midrule
\multicolumn{6}{l}{\emph{By difficulty level}} \\
 & Level 1 & Level 2 & Level 3 & Level 4 & Level 5 \\
Tasks & 1{,}644 & 2{,}202 & 2{,}094 & 1{,}886 & 2{,}083 \\
Success (\%) & 90.5 & 88.1 & 83.4 & 74.3 & 56.3 \\
\bottomrule
\end{tabular}
\end{center}
\end{table}

\paragraph{By suite and difficulty.}
Table~\ref{tab:liberoplus-breakdown} shows that the two more semantic suites degrade most under perturbation: relative to their unperturbed counterparts in Table~\ref{tab:libero}, Goal loses 23.0 points (98.0 to 75.0) and Long 21.9 (95.0 to 73.1), versus 16.4 for Spatial and 14.9 for Object.
Across the benchmark's difficulty stratification, success decays monotonically from 90.5\% at Level~1 to 56.3\% at Level~5.
This graded degradation, rather than a bimodal solved/unsolved split, indicates that the 78.4\% total is not carried by the easy strata alone.

\begin{table}[t]
\caption{\textbf{The Sensor Noise column of Table~\ref{tab:liberoplus}, split by corruption type and severity tier.} Types are encoded in the benchmark's task naming. Success rates in \%.}
\label{tab:liberoplus-noise}
\begin{center}
\small
\begin{tabular}{lcccc}
\toprule
Corruption type & Overall & Low & Mid & High \\
\midrule
Motion blur & 87.8 & 100 & 95 & 76 \\
Glass blur & 85.7 & 93 & 89 & 79 \\
Gaussian blur & 69.2 & 95 & 78 & 47 \\
Zoom blur & 62.5 & 96 & 77 & 44 \\
Fog & 60.7 & 93 & 80 & 39 \\
\bottomrule
\end{tabular}
\end{center}
\end{table}

\paragraph{What the noise split reveals.}
The 74.2\% Sensor Noise column is not a uniform weakness.
Split by corruption type (Table~\ref{tab:liberoplus-noise}), the dividing line is not static versus dynamic corruption but whether the corruption preserves the geometric evidence in the visual stream.
Motion blur and glass blur smear textures and edges while leaving the global scene layout intact and are largely absorbed (87.8\% and 85.7\%).
Zoom blur superimposes rescaled copies that shift objects' apparent positions.
Fog injects a per-frame random low-frequency luminance field that makes the history window temporally inconsistent.
Gaussian blur at high severity erases object boundaries outright.
Success collapses exactly on these three (62.5\%, 60.7\% and 69.2\% overall, falling to 44\%, 39\% and 47\% at the highest severity).
All five types degrade monotonically with severity, indicating graded loss of evidence rather than a threshold artifact.
SimpleMemVLA leads on Camera, Light and Background perturbations, while performance remains weaker on Language and several sensor-noise conditions.

\section{Additional Analysis}
\label{app:analysis}

\subsection{The Gains Track Memory Load}
\label{sec:analysis-gains}

If the gains came from a generically stronger backbone or better low-level control, they would be flat across memory loads. Instead the advantage grows along each suite's own memory axis (Tables~\ref{tab:rmbench}, \ref{tab:robomme} and~\ref{tab:mikasa}). On RMBench the margin over Mem-0 widens from +38.8 on single-memory tasks to +68.5 on multi-memory ones. On RoboMME it grows from +24.7 on Counting, whose periodic cues stay partly visible, to +44.5 on Reference, whose evidence is entirely off-frame. On MIKASA-Robo the lead over the best prior VLA stretches from 12 to 39 points as RememberColor moves from 3 to 9 candidates. Where memory is not the bottleneck the margin does not reverse but vanishes, since on LIBERO the same recipe is simply on par with the best reactive VLAs at 97.5\%. This dose--response pattern ties the gains to memory itself.

The same lens delimits the claim. SimpleMemVLA's residual failures are consistently not retention failures: they concentrate in precision control (Insert Peg at 46\%, where even the perfect-perception oracle reaches only 15.6\%, Table~\ref{tab:robomme-full}), in fine-grained re-identification (Observe\&Pickup 65\%, RememberColor-5 58\%, Video Repick 64\%) and in perturbations that rewrite the instruction or attack positional evidence itself (Appendix~\ref{app:liberoplus-detail}). Feeding the policy its own past removes the need for memory machinery. It does not substitute for perception or control.

\subsection{Why the Machinery Falls Short}
\label{sec:analysis-machinery}

The controlled comparison of Section~\ref{sec:controlled} leaves a puzzle: the prior designs are far more elaborate than ours, yet they lose by wide margins. The following comparisons and interventions provide complementary evidence about the historical information needed at decision time.

The first is a controlled comparison that the RoboMME benchmark itself provides, since its fourteen memory variants share one $\pi_{0.5}$ backbone and differ only in the mechanism. The pool covers all three families of Figure~\ref{fig:overall}, with external stores in both symbolic and retrieval form: symbolic scene-graph stores reach 11.6 to 32.7\%, the retrieval-based MemER 42.4\%, learned compression 30.7 to 44.5\% and recurrent state 18.1 to 22.3\%, against 88.3\% for attending to the timestamped stream itself (Table~\ref{tab:robomme}). Our own rebuilds compare alternative memory interfaces on a shared backbone and training stack, reaching 20.6--31.5\% against 88.3\% for native context.
These results characterize the evaluated implementations under the training protocols described in Appendix~\ref{app:impl}. More machinery does not help: the recurrent family's best variant stays at 22.3\% and the lowest scores in the table belong to symbolic stores with a VLM in the write loop. The ordering holds beyond RoboMME as well: the strongest prior memory VLA reaches 44.4\% on MIKASA-Robo where the raw stream reaches 74.0\% and 73.1 against 78.4\% zero-shot on LIBERO-Plus (Tables~\ref{tab:mikasa} and~\ref{tab:liberoplus}).

The second experiment removes write quality as the explanation. GroundSG handed ground-truth VLM outputs is a write-time abstraction executed perfectly, with no perception error and no capacity limit, yet it reaches 84.1\%, below the 88.3\% of the raw stream. The ceiling of this abstract-then-store pipeline sits below the input it abstracts, because what the write discards before the need is known stays missing at read time. Running the same pipeline with real perception loses another fifty points (32.7\%) and the strongest deployable baselines collapse exactly where the needed evidence is a specific past percept: MemER falls to 38.0 and 53.2\% on Reference and Permanence, Mem-0 from 52.8 to 28.5\% as the number of facts grows and MemoryVLA++ from 97\% on ShellGame to 16\% on RC-9, where one cue must stay distinct from nine alternatives.

The third line of evidence comes from the deployment edits of Sections~\ref{sec:analysis-read} and~\ref{sec:ablations}, which name what each abstraction discards. Shuffling frames zeroes both probe tasks and a retrieved set of keyframes is exactly an unordered set. 
The success cliff at the evidence age shows the read needs the stale frames that recency-biased compression evicts first. 
Zeroed timestamps halve counting and neither summaries nor recurrent states keep a time ruler. 
The graying interventions show the read consumes the percept itself, which no symbolic entry can reproduce. 
The recurrent rebuild compresses visual content, temporal order, timestamps and older evidence into a fixed sixteen-token state.
It reaches 20.6\% overall and is strongest on Counting at 31.0\%, consistent with partial retention of running totals but not with reliable preservation of the perceptual and temporal evidence required across the suite.
The keyframe bank discards only the order and the timestamps while keeping the percepts, so its 31.5\% is both higher and selective in the way this list predicts, since the tasks it survives are the ones whose answer needs no time ruler. 
The compressor squeezes the percepts through a fixed budget, the very channel the graying interventions single out, and its 22.6\% is selective in the complementary direction, keeping the tallies a summary can hold while the percept-dependent tasks the bank kept raw collapse. 
The one edit that never hurts is the control that keeps the evidence frames ordered and intact while removing others, so selection as such is not the problem. Discarding before the need is known is.

Nor does the dilemma leave a lossless way out. A write format that keeps the order, the timestamps, the stale frames and the percepts themselves has stopped abstracting, because up to the sampling rate it is the stream. A store that keeps everything and defers every choice to read time escapes the dilemma, but read-time selection over an intact past is what attention over the window already does, with no store to maintain. Attention over the raw stream wins by refusing the choice every tested family made: it defers selection to read time, when the need has arrived.

\section{Bounded-Cost Streaming with Sliding-Window Attention}
\label{app:swa}

Streaming Inference with Prefix Prefill keeps decision latency close to the single-frame baseline for minute-scale windows.
Its cache size depends on the retained context length and remains bounded under the standard 60\,s cap.
The SWA variant maintains a persistent stream with block-wise cache eviction.

\paragraph{Window the softmax layers, and only them.}
The backbone interleaves two kinds of sequence mixing: linear-attention layers, whose recurrent state is constant-size and summarizes the entire stream by construction, and softmax-attention layers, the only place where per-token state accumulates (24 and 8 of the 32 layers, respectively).
SWA therefore touches nothing but the 8 softmax layers.
During training and at deployment alike, each of their queries attends to at most the most recent $L$ tokens; the linear-attention layers are left untouched and keep integrating every token since the first frame of the episode.
The window is sized in whole history units, where one unit is the token block of one temporal patch: its plaintext timestamp (6--8 tokens), the vision delimiters and 80 video tokens, 88--90 tokens in all.
We set the window to $M=60$ units, exactly the 60\,s memory span the recipe already deploys on RMBench (Table~\ref{tab:impl}), plus the decision suffix and the generation budget, giving $L=5{,}888$ tokens with margin.
Inside the window nothing changes: selection over the explicit past remains read-time attention, the thesis of Appendix~\ref{sec:analysis-machinery}.
Evidence older than the window survives only through the linear-attention state, the same finite-span caveat the windowed recipe always had, now with a compressed residue beyond it instead of nothing.

\paragraph{Timestamps must be absolute.}
A streamed token may never change after it is emitted, otherwise its cached key--value entries are invalid the moment the window slides.
The window-relative timestamps of the deployed recipe violate exactly this: one decision later the same physical frame carries a label smaller by one execution interval, so every unit would need re-encoding at every step.
The SWA variant therefore switches the timestamp convention from window-relative to episode-absolute, computed from each frame's index on the episode's sampling grid through one function shared by training and deployment.
Under absolute time a unit's tokens are immutable, its cache entries are written once and reused until evicted and positions continue monotonically without ever being renumbered.

\paragraph{The conditioning interface needs no change.}
The action head conditions only on the generated answer span and the state token (Equation~\ref{eq:cond}), which sit at the tail of the sequence and hence always inside the window, so the train/deploy isomorphism of Section~\ref{sec:method-head} carries over unchanged.

\begin{figure}[t]
\centering
% Source: figures/src/fig_swa_method.tex (standalone TikZ; rebuild with tectonic/pdflatex).
\includegraphics[width=\textwidth]{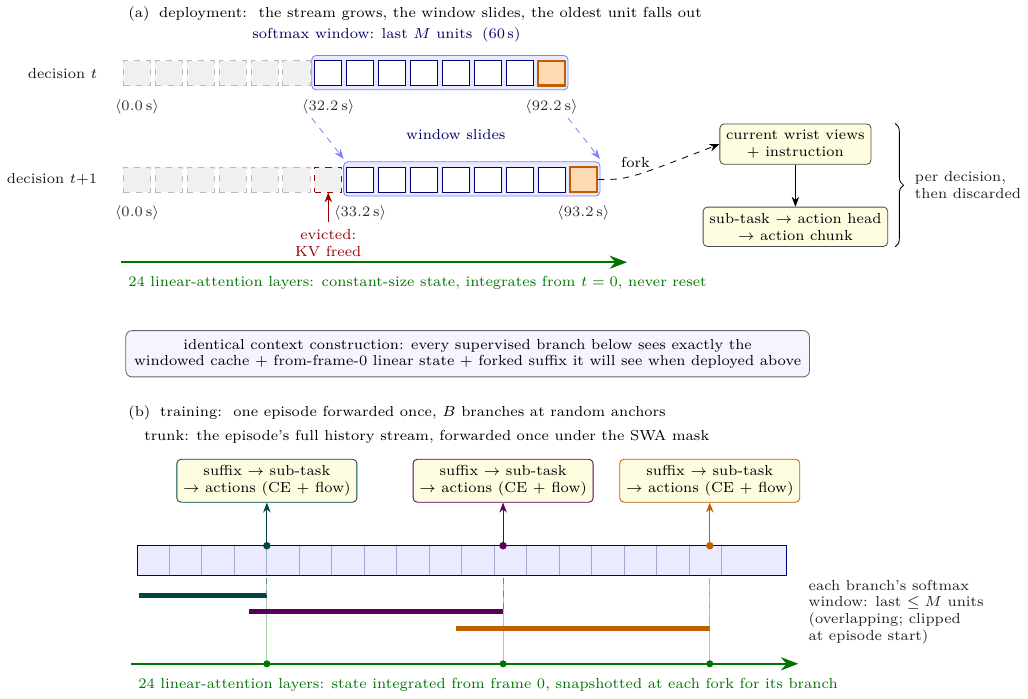}
\caption{\textbf{The SWA variant: one context construction, trained and deployed.} \textbf{(a)} Two consecutive streaming decisions: each appends one unit (orange) to the persistent stream, the softmax window slides by one unit, and the unit leaving it (red) has its cache rows freed; the linear-attention state (green) integrates from $t=0$ and is never reset. \textbf{(b)} Training forwards each episode once as a trunk and supervises $B$ branches at random anchors, each attending to its last $\le M$ units and resuming the linear state snapshotted at its fork — exactly the deployed construction. Details in the accompanying text.}
\label{fig:swa-method}
\end{figure}

\paragraph{Deployment: append one unit, evict one unit.}
Figure~\ref{fig:swa-method}a shows one decision.
The persistent stream holds only the history video channel.
When a decision fires, the newest temporal patch is encoded into one unit and appended (positions continue from the previous tail), softmax-cache rows older than the window are dropped in whole units and the linear-attention states are simply kept.
The decision suffix, current wrist views, instruction and chat scaffold, is forked off the stream state, the sub-task is decoded, the action head conditions on its span and the fork is discarded.
Eviction is exact rather than approximate: a dropped row is outside the window of every query that will ever be computed, so the streamed decision reproduces the full-prompt SWA decision token for token, which we verify at integer exactness for token and position ids and at kernel-noise level for activations.
Per decision the variant therefore encodes one temporal patch, prefills one unit and decodes one short answer against a bounded cache: compute and memory are both independent of how long the robot has been running.

\paragraph{Training with episode-level packing.}
Training must expose the model to the same context construction, and the subtle point is the linear-attention layers.
Ordinary per-anchor training renders each anchor a finite window and integrates the recurrent state from that window's first frame, while the streaming deployment integrates it from the episode's first frame, a train/deploy mismatch in exactly the pathway SWA makes load-bearing.
SWA training therefore packs samples at the episode level (Figure~\ref{fig:swa-method}b).
One sample is an (episode, phase) pair, where a phase is one residue of the temporal-patch grid, so that every anchor of the sample forks on a unit boundary of one shared trunk rendering.
The trunk, the episode's full history stream, is forwarded once with the SWA mask on the softmax layers and the linear-attention state running from frame $0$; $B=8$ anchors drawn from the phase's chain fork off it, each resuming from the linear-attention states snapshotted at its fork and attending to the last $M$ units of trunk cache, and each pays the standard losses of Section~\ref{sec:method-head}.
The packed pass is not an approximation: token and position ids match the per-anchor construction exactly and the supervised span's hidden states match within bf16 kernel noise, while the shared trunk cuts per-anchor compute several-fold.
Random anchor draws with repeats cover every anchor near-uniformly across epochs.
Relative to the recipe of Section~\ref{sec:method}, the SWA variant changes only the attention mask, the timestamp convention and the packing.

\begin{figure}[t]
\centering
% Source: figures/src/fig_swa_cost.py (matplotlib; reads the single-GPU bench JSON
% from experiments/bench_stream_vs_recompute.py in the memVLA repo).
\includegraphics[width=\textwidth]{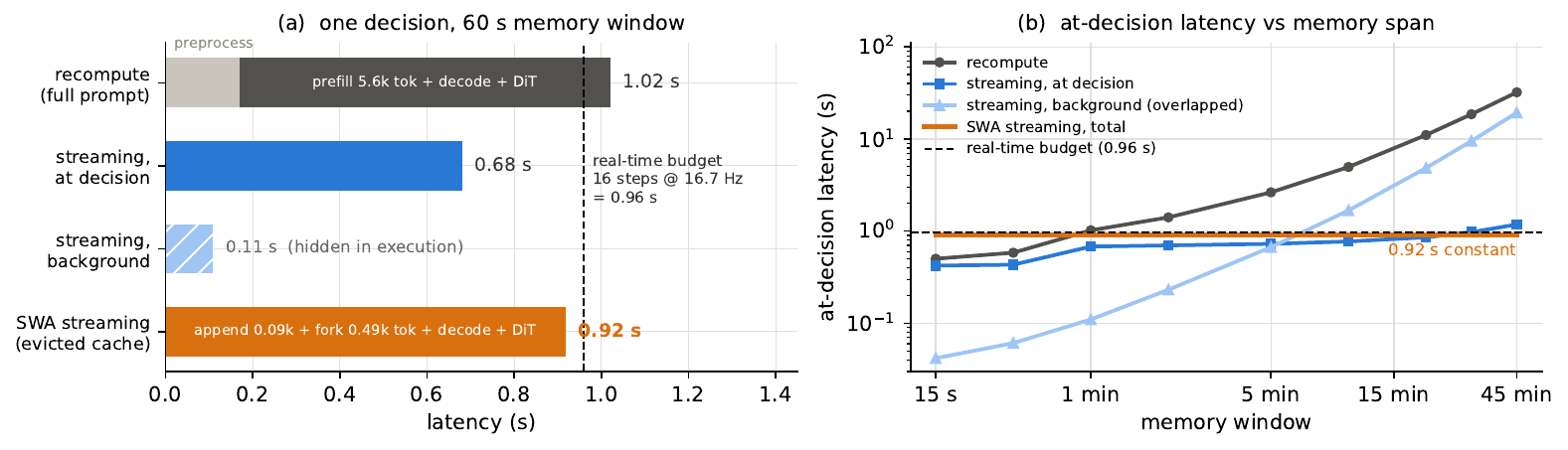}
\caption{\textbf{Figure~\ref{fig:latency}, extended with the SWA streaming variant.}
Measurements use one H100, bf16 and batch size 1, with the setup of Figure~\ref{fig:latency}.
\textbf{(a)} The SWA decision takes 0.92\,s, within the 0.96\,s execution interval.
\textbf{(b)} The SWA variant uses a fixed $L=5{,}888$-token window to bound its resident cache.}
\label{fig:swa-cost}
\end{figure}

\paragraph{Cost.}
Figure~\ref{fig:swa-cost} places the SWA variant on the axes of Figure~\ref{fig:latency}.
The long-input experiment measures the cost of increasing the retained context length; its 45\,min input approaches the backbone's 262k-token limit.
Standard deployment instead caps the history window at 60\,s.
With a fixed $L=5{,}888$-token attention window, the SWA variant bounds the resident cache and the number of tokens processed per decision.
Its measured median decision latency is 0.92\,s over 153 consecutive decisions.

\begin{table}[t]
\centering
\caption{\textbf{Closed-loop success of the SWA streaming deployment}, on the RMBench protocol of Table~\ref{tab:rmbench} (100 held-out seeds per task).}
\label{tab:swa-results}
\resizebox{0.99\linewidth}{!}{
\begin{tabular}{lccccccccccccc}
\toprule
 & \multicolumn{6}{c}{Single-memory $M(1)$} & \multicolumn{6}{c}{Multi-memory $M(n)$} & \\
\cmidrule(lr){2-7}\cmidrule(lr){8-13}
Method & \rotatebox{90}{Obs\&PU} & \rotatebox{90}{Rearr.} & \rotatebox{90}{PutBack} & \rotatebox{90}{SwapB} & \rotatebox{90}{SwapT} & \emph{Avg} & \rotatebox{90}{Battery} & \rotatebox{90}{Rank.} & \rotatebox{90}{Cover} & \rotatebox{90}{Press} & \rotatebox{90}{Place-Mat$^{\dagger}$} & \emph{Avg} & Overall \\
\midrule
SimpleMemVLA (Table~\ref{tab:rmbench}) & 65 & 100 & 100 & 100 & 93 & 91.6 & 90 & 100 & 98 & 100 & 100 & 97.0 & 94.0 \\
\rowcolor{orange!15}
$+$ SWA streaming (evicted cache) & 56 & 100 & 100 & 100 & 93 & 89.8 & 71 & 100 & 99 & 100 & 99 & 92.5 & 91.0 \\
\bottomrule
\end{tabular}
}
\end{table}

\paragraph{Results.}
Table~\ref{tab:swa-results} evaluates the SWA streaming deployment on the full RMBench protocol.
Seven of the nine scored tasks stay within one point of the Table~\ref{tab:rmbench} deployment (five at 100\%), and the overall score is 91.0 vs.\ 94.0.
The remaining difference is concentrated in the two hardest memory tasks (Obs\&PU and Battery) and reflects training under the SWA regime rather than the streamed inference itself, which reproduces the policy's full-prompt decisions exactly (verified at integer exactness for token and position ids and at kernel-noise level for activations, as argued above).
The variant thus trades a small amount of accuracy on the hardest memory tasks for a deployment whose per-decision compute and memory are constant in episode length.

\end{document}